# Introduction to ROSS: A New Representational Scheme

**Glenn R. Hofford**
Software Engineering Concepts, Inc.
glennhofford(at)gmail.com

## Abstract

Representation is a cross-discipline topic that includes knowledge representation from the field of artificial intelligence, meaning representation from the field of natural language understanding, and structured information from information processing. The field of logic is replete with methods for representation and reasoning. In the areas of AI knowledge representation and reasoning, leading approaches include first order logic, description logic, lambda calculus, semantic networks, frame-based approaches and many others. Meaning representation techniques from the field of natural language understanding include the above in addition to approaches like conceptual dependency.

ROSS ("Representation, Ontology, Structure, Star") is introduced as a new method for representation that emphasizes representational constructs for physical structure. The ROSS approach starts with a normal form that is self-consciously less-expressive than logic, and builds from the bottom-up to achieve the expressiveness that is needed for a domain. Information that uses the normal form exists within an analogical frame of reference. What is achieved is a greater degree of structure in the representations that use ROSS when compared with representations that are typically created using formal approaches such as FOL. The ROSS normal form involves a set of modeling guidelines and underlying ontological commitments that involve naive or intuitionistic assumptions.

ROSS can be used in several ways: it is a representational method for knowledge bases that contain class definitions; it is a method for representational artifacts that contain fact-like constructs; and, it provides a foundation for representations that support various forms of inference. The first two uses are described and a brief introduction to how ROSS supports reasoning techniques is presented. The principles of ROSS have been used to create an expert system for diagnosis and a proof of concept natural language understanding system for story comprehension; these principles are briefly described.

## 1 Introduction

ROSS is introduced as a representational approach that contains unique features that assist with a number of important representation objectives. ROSS is an acronym composed of the initial letters from the words "Representation", "Ontology", "Structure", and "Star language". The Star language is also introduced in this paper. ROSS provides an infrastructure and a set of implicit guidelines that help with the following tasks:

- Creation of *knowledge bases* that have well-organized entity and behavior classes.
- Generation of representation artifacts (containing *instances* of classes) that are highly structured based on the use of a *normal form* for spatial/temporal location-related attributes.

ROSS accomplishes two things that relate to the use of the term "structure"; the distinction between these two aspects is as follows:

- It represents the *physical structure* of a problem domain or domain of discourse in a way that is rich and deep. How this is specifically done is explained.
- The representational *knowledge base* or *fact artifact* itself is highly structured. Two specific methods are presented: Star language definition documents and an XML-based format for instance models. (An instance model is a meaning representation artifact that is generated by the semantic processing of a natural language understanding program).

It will be shown that these two features of ROSS are complementary with each other.

ROSS is also a representational platform for a variety of techniques for automated inference. The set of ROSS features for inference is briefly introduced.

### 1.1 Context Within Information Processing, AI and NLP

The field of information processing requires methods that support and facilitate structured storage (via key-based indexing), that in turn supports query and analysis. The DBMS field has well-established conventions and methods for these tasks. ROSS is useful for creating structured information artifacts that go beyond DBMS in the complexity of the subject matter or problem domain.

The field of symbol-based AI knowledge representation and reasoning requires techniques that allow automation of inference in its various forms. Formal methods for knowledge representation are necessary to facilitate reasoning. *Knowledge representation artifacts* can be classified with respect to the types of information they represent: artifacts that include abstract definitions, e.g. of classes, and rule-like constructs are referred to as *knowledge bases* or *rule bases*. Artifacts that include fact-like constructs, whether these are past facts, goals, plans, predicted states or other fact-like items are referred to herein as *fact repositories* or *transcripts*.

In the field of natural language processing (NLP) and natural language understanding (NLU), a *meaning representation* is a cohesive representational artifact that is the output of a process of natural language understanding or comprehension as this process is applied to a sentence, fragment, or document of human natural language text. Existing meaning representation approaches include first order logics, semantic networks, and frame-based approaches (Jurafsky and Martin, 2009). Other approaches include lambda calculus, description logic and conceptual graphs. The main objective of such natural language understanding meaning representation techniques and languages is that of representing the original information in a structured way – this enables further NLP objectives that include tasks as varied as summarization, named entity recognition, and relationship detection. Question answering systems require the transformation of questions into a structured form. NLU can also benefit from structured meaning representations that are a platform for inference about semantics and context: this allows for enrichment of the parsing and semantic processing steps. Semantic processing within NLU also involves the use of knowledge bases that contain classes of entities and behaviors; ROSS provides an infrastructure for this requirement.

ROSS is a representational method that is *physical symbol-based*[1]. Generally speaking it is based on the tenets of the *Knowledge Representation Hypothesis* [2]. Knowledge (definitions, facts, rules, etc.) is represented *declaratively* rather than *procedurally*. ROSS can be used as a representation and reasoning technique that assists human cognition, and, it is capable of automation. As a method that can be automated, it provides a representational infrastructure that allows for effective inference in an independent mechanistic manner.

### 1.2 Ontological Basis

The ontology of ROSS is an "operational ontology" that has been worked out for the purpose of representation. ROSS starts with a seemingly simple ontological concept and works out the implications and details consistently. The concept addresses the question "*for the purpose of constructing structured representational knowledge bases and artifacts*, what *exists* in the physical or external world, that is, the world (whether real or hypothetical) that is represented?" The answer is: unit-sized location entities[3] that do not move; these entities exist within a four-dimensional (4D) world that can be understood as consisting of Euclidean space with the added dimension of time. Cartesian coordinates are used with the important restriction that dimensions are only represented using integers, not real numbers (this restriction is necessitated by the requirement of indivisibility and is explained in the section on the ontology of ROSS). The ontological foundations thus share similarities with those of the fields of naive physics and commonsense representation and reasoning[4].

The ontological concept involves the idea that a unit-sized location entity is an entity that has both spatial size and temporal duration. The representation of other entities, aspects, or characteristics of the represented world from the universe of discourse is accomplished using a variety of representational constructs for aggregation or composition. For instance, larger entities are represented using a mechanism for the specification of structure within a hierarchy, and motion is broken down into a sequence of states, each of which is represented using attributes that specify the static features of a single-time-point entity.

---

[1] Cf. Newell and Simon (1976) for a general definition of a *physical symbol system*.

[2] The *Knowledge Representation Hypothesis* is described in Smith (1982).

[3] The term "entity" is used somewhat imprecisely throughout this paper – in general it refers to a fixed location in space and time. The difference between "entity" and "object" will be explained in what follows.

[4] Cf. Mueller (2006) - the discrete event calculus as an example of an integer-based system.

The ontology of ROSS consists of these and several other ontological constraints and requirements, which collectively provide a basis for the creation of a set of knowledge base *definitions*. In the area of NLU for instance, these definitions can be used to assist or support the (human or automated) task of the creation of *fact transcripts*, or *instance models* that represent the subject matter of natural language textual input.

### 1.3 Epistemology

ROSS in its current form does not directly address the epistemological (i.e. *belief-related*) aspects of representation. This is primarily due to the emphasis that is placed on the structural aspects of representation. An implicit assumption is involved wherein a ROSS fact repository is an artifact that represents a set of true facts about a situation (the degree of belief involves full certainty). Where a situation involves agents that themselves represent and communicate information, the agents' mental states and the communicative aspects are represented using the same fundamental approach that is used for representing entities and aspects of commonsense domains. (However, ROSS does have features for representation that involves a high level of abstraction; these features are well-suited for representations of representation, information, mental states, communication, etc.).

### 1.4 Is ROSS a Logic?

ROSS is a method that fits the definition of "representational scheme" (Hayes, 1974). ROSS has a set of syntactic rules and semantics that are based on a set of ontological assumptions and on an ontological framework that builds on these assumptions. ROSS shares many features with representational schemes that are referred to as "logics" such as first order logic and description logic. ROSS provides a foundation for reasoning, or inference; however, unlike most logics, the techniques for reasoning that use ROSS are *loosely coupled* with the representational system of ROSS. This loose coupling allows ROSS to support a wide range of methods for inference. However, some of the normally expected elements for a logic are not currently present in ROSS, such as a set of axioms; therefore the term "representational scheme", rather than "logic", has been deemed as appropriate for ROSS.

Practitioners who are familiar with using logic for the tasks that are addressed by ROSS may find that the concepts of ROSS seem somewhat cumbersome. ROSS is a paradigm shift from logic, and using it requires additional effort in some areas such as the tasks of modeling to create supporting definitions and to create intricate classes. In addition, ROSS imposes stringent restrictions on the modeling task; these may at first seem unnecessary. This paper attempts to show the benefits that result from the effort that is required by the technique.

### 1.5 Comprehensiveness of Representation

A requirement for all structured representation approaches is that of comprehensiveness or completeness – the meaning representation language or logic must be capable of full coverage of the information for a problem domain or from a domain of discourse (for NLU, this is the domain of the information represented by the input natural language text). ROSS provides a framework that allows for comprehensive representation, both in knowledge bases and in fact-related artifacts.

A sometimes-overlooked or misunderstood objective for a representation is that of *organization*. It is often mistakenly assumed that the way to achieve highly structured representational artifacts involves using highly refined, elegant and concise techniques (e.g. FOL, lambda calculus). These approaches are appropriate for some domains, such as the task of representing numerical concepts and computational processes. However what has been overlooked is that a method may be highly formalized and mathematically precise, yet at the same time it may be weak for practical use for domains and representational uses outside the scope to which it has been applied. In contrast, the premise of ROSS is that a large field of domains has a rich structure that is readily organized in ways that have not been accomplished with other techniques.[5]

### 1.6 A Non-Objective: Conciseness of Expression

ROSS knowledge bases and representational artifacts are perhaps less *concise* than those of other logics or other schemes. ROSS definitional constructs (attribute value sets, attribute types, classes, etc.) are somewhat elaborate. The goal of representation with ROSS is a high level of organization, not necessarily conciseness of its definitions and expressions. The task of modeling in order to create a set of classes and instances that are appropriately comprehensive and complex - even for some seemingly-simple assertions - necessitates a rich, complex and sophisticated representational approach.[6]

[5] Expressiveness, e.g. of FOL, as that which involves universal quantification, negation, disjunction and nested assertions is not viewed here as identical to comprehensiveness and is treated separately in this paper.

## 1.7 Background

Throughout the 60-plus years of their respective histories, the fields of AI knowledge representation and reasoning and NLP natural language understanding have involved quests for representational methods that are capable of representing problem domains or the subject matter from domains of discourse in ways that are suitably rich and deep. A deep representation would be one that is comprehensive and that adequately captures the structure of the represented domain. The AI tasks of knowledge representation and commonsense reasoning have been addressed by practitioners using symbolic approaches as early as the 1960s and 1970s. This work flourished during that period and extended into the 1980s, but symbolic approaches ran into difficulties and did not achieve the successes that had been anticipated.

An example is the SHRDLU system of Terry Winograd, which focused on commonsense reasoning about simple domains and question-answering (Winograd, 1971).

Within the field of natural language understanding, (Sowa, 2006) describes the shift that took place during the 1980's on the part of Terry Winograd and others:

> *Terry Winograd, for example, called his first book Understanding Natural Language (1972) and his second book Language as a Cognitive Process: Volume I, Syntax (1983). But he abandoned the projected second volume on semantics when he realized that no existing semantic theory could explain how anyone, human or computer, could understand language. With his third book, coauthored with the philosopher Fernando Flores, Winograd (1986) switched to his later work on the design of human-computer interfaces. Winograd's shift in priorities is typical of much of the AI research over the past twenty years. Instead of language understanding, many people have turned to the simpler problems of text mining, information retrieval, and designing user interfaces.*

The ambitious systems of the 1970s and 1980s were viewed either as overly-complex, cumbersome, and overly domain-specific. Some of these projects perhaps raised more questions than they answered. Subsequently, particularly in the NLP field, statistical approaches came to the fore and have tended to dominate the landscape.

The question of why the symbolic approaches failed - or only achieved partial success - has continued to perplex those who believe that important answers are yet to be discovered that involve symbolic approaches. The author's view is that the ROSS approach provides a foundational set of principles and a representational scheme that contain important answers to these questions.

## 1.8 Comparisons

This paper compares ROSS against first order logic (FOL) and description logic; a few comparisons are also made against frames. In contrast to FOL, ROSS is based upon a set of *explicit assumptions* about the nature of the reality that is represented (the *represented world*) in order to achieve a greater degree of organization than is typically achieved using FOL. ROSS also includes important *modeling restrictions* that constrain the task of creating definitions that model the world or domain.

In contrast to description logic, ROSS is not built on a syllogistic categorical approach. ROSS classes are not representations of *sets*, a ROSS class is a mechanism for storing information about instances that get instantiated. Inheritance is optional in ROSS.

---

[6] Nash (2013): a blog post entitled "Make Things As Simple As Possible, But Not Simpler" emphasizes the danger of oversimplification in technology. ROSS is aligned with the premise of the so-called *Einstein's Razor*: "It can scarcely be denied that the supreme goal of all theory is to make the irreducible basic elements as simple and as few as possible without having to surrender the adequate representation of a single datum of experience." (courtesy: http://en.wikiquote.org/wiki/Albert_Einstein) This can be summarized as "Make things as simple as possible, *but not simpler*." Contrast with Occam's Razor: "plurality must never be posited without necessity", which may in some cases be used as a justification for *oversimplification*.

ROSS shares some similarities with frames[7], but it has a much more elaborate infrastructure than frames for representing physical structure. For readers who are accustomed to the *frames* terminology, a ROSS *attribute type* is the rough equivalent of a frame *slot*, and a ROSS *attribute value* is the rough equivalent of a frame *filler*.

## 2 Core Concepts

The following sections describe the core concepts of the ROSS approach.

### 2.1 The Physical Structure of the Domain is Richly Represented

Can traditional logic represent the complex physical structure of a person? The answer in theory is yes; however in practice traditional logics provide little in the way of guidelines for accomplishing representational tasks like this. For instance, given a hypothetical "person" class, how is the set of spatial relationships between the overall person and the sub-parts (components) such as person head and person body represented? How is physical size represented?

The representation of part to sub-part ("PartOf") relationships is a foundational and well-studied task within AI knowledge representation [8]. A number of AI representational schemes, including frames and KL-ONE do represent physical structure and the PartOf relationship. However these techniques are limited[9]. In order to facilitate *completeness* of representation, firstly, a representational scheme should support the following features; these go well beyond predication of the existence of a "PartOf" relationship:

- attributes that can specify the location of a component part in relation to its parent object.
- attributes that can specify the spatial orientation of a component in relation to its parent object.
- attributes that can specify the physical dimensions of a component (its size, or extent) using a coordinate scheme that is defined in terms of its parent object.

Secondly, this should be done in a way that is *flexible*, so that when instances get instantiated based on an entity class, they are created with a basic structure, but only the specifics that are known are specified.

#### 2.1.1 Object Frame Class

ROSS introduces two important representational constructs that handle structure. The first of these is called the *object frame class*.[10] An object frame class represents a specific physical location within a four-dimensional space-time frame of reference. The object frame class is a class that is used for the creation of object frame *instances*: an object frame instance is simply a location that is fixed in space and time. (Note: an object frame class is sometimes referred to as a "location entity" class – these terms are synonymous). An object frame class or instance can be (spatially) "unit-sized" or "aggregate". An aggregate object frame instance is somewhat analogous to a rectangular wire frame. (The upcoming Ontology section describes the background of these concepts in greater detail).

#### 2.1.2 Dimension System

The second representational construct is called the *dimension system type*, or *dimension system*. A dimension system is a group of closely-related *dimensions*. A dimension, or *dimensional attribute*, is like a single coordinate within a coordinate system such as the Cartesian coordinate system (e.g. the value of an x coordinate is a dimension). A dimension system includes a set of such attributes that must be used together in expressions that specify the location of a physical object. Dimension systems are used within object frame classes to provide a way of specifying locational[11] attributes of embedded component object frame classes. The component object frame class uses a special "RelationshipToParent" section, in which the dimension system items are used in order to specify the location and orientation of the component in relation to the parent, and in order to specify the dimensions (size or extent) of the component.

**Figure 1** shows the structure of the ROSS object frame class that represents the structure of a person, called “PersonObjectFrameClass”. Note that the detail for each of the two dimension systems of the PersonObjectFrameClass is not shown.

[7] Minsky (1981) originated the *frame* concept; Minsky (1986) contains further elaborations on frames.

[8] Sometimes referred to as *mereology*.

[9] Minsky (1981), Brachman and Schmolze (1985)

[10] "Frame" here is not equivalent to the AI frames; in the context of "object frame class", it denotes a structural aspect.

[11] The term "locational" is used by the author to refer to the location aspects of an entity; these can be spatial or temporal aspects.

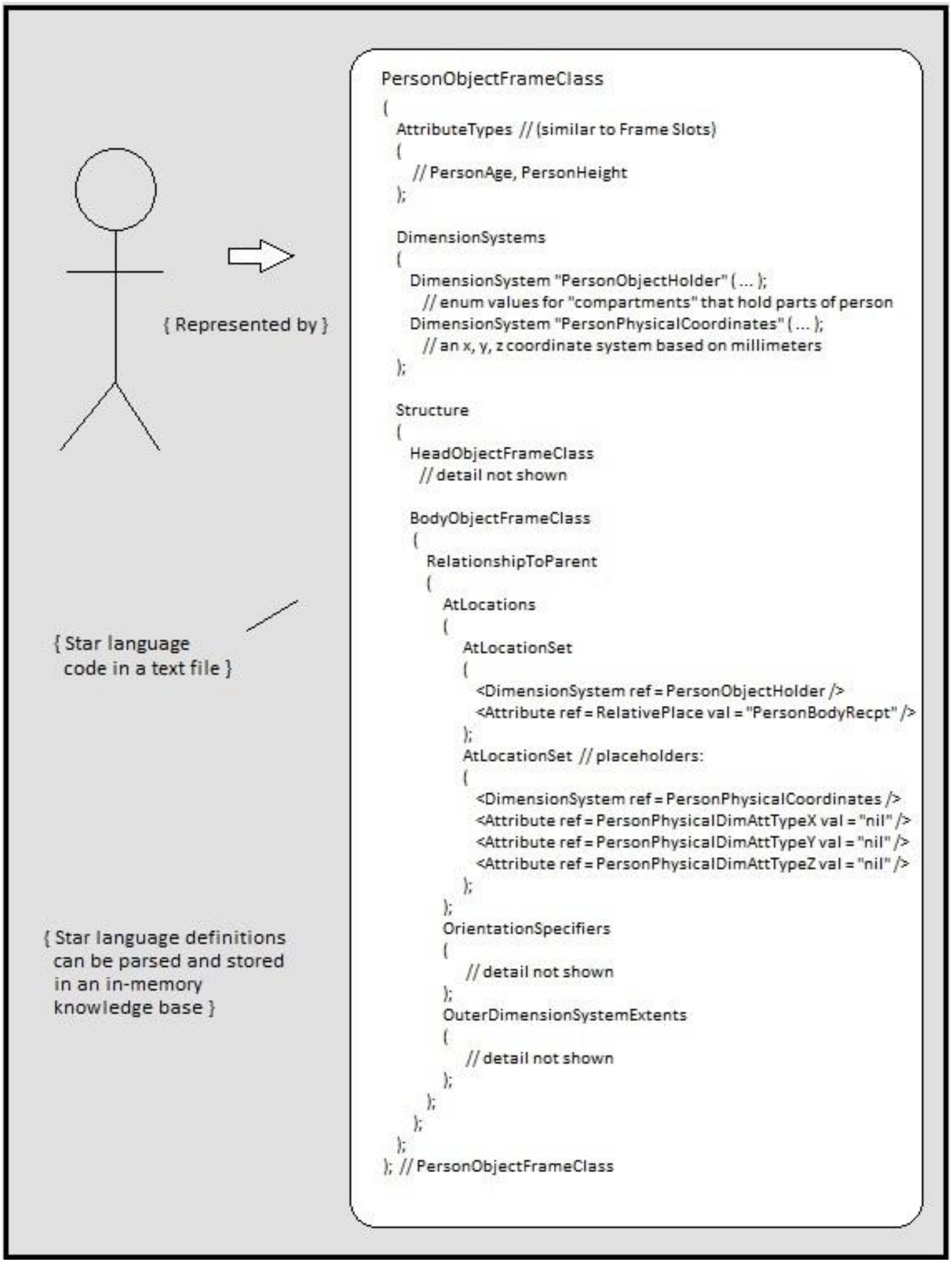

**Figure 1: The Structure of an Object Frame Class**

The diagram provides an overview of how an object frame class represents structure. Because this is a class, the actual attribute values for a number of items are not filled in: they are designated as “nil”. When an instance is instantiated from the class, if specification information is available it will be used to fill in these values.

Object frame classes and dimension systems facilitate the creation of representations that represent *dimensional structure*. This dimensional structure is usually defined by the use of spatial and temporal dimensions; however other non-specific dimensional attributes can be used, such as enumerated values. An example would be a dimension system that has a locational attribute with values of PersonHeadReceptacle and PersonBodyReceptacle. (A *receptacle* is an attribute value that designates a location that is distinct from other receptacles that are part of the enumerated list).

This approach provides an infrastructure for what is referred to as “primary information”:

propositional information that is tied into a frame of reference. This eliminates the need for a multiplicity of assertions that represents not only essential (qualitative) facts but the relational "place" of the objects of those facts (such as is typically done with FOL). (As will be seen, relationships between location entities are handled by ROSS; such relational information is part of the category of secondary or derivative information).

### 2.2 Definitions of classes

Besides class definition statements that define object frame classes (aka "entity classes") ROSS contains other definitional constructs such as attribute types, attribute value sets, dimension systems and behavior classes. These are rich, sophisticated definitions that provide context for propositional expressions which are used in specific representations. Consequently, the instances that are instantiated from the object frame classes have a rich set of structural, attributive, relational and behavioral attributes.

ROSS classes can be compared to those of the description logics, where the *concept* can[12] represent a *class*. Description logics provide a way to define a unary predicate (concept) in terms of constituent features. Like description logic classes, ROSS classes can contain attributive and relational information (e.g. the class of gold coins has the attribute of having gold material composition). However ROSS classes also represent "type" information involving dimension systems (i.e. dimension system "types"), attribute types and relationship types. In addition, structural information in a ROSS object frame class can be viewed as "type" information), and object frame classes have lists of associated potential behaviors; this is another form of type information.

In contrast to some representational schemes and logics, a ROSS class is not equivalent to a mathematical set. A ROSS class is a mechanism for the aggregation of features.

**Figure 2** is an overview of the supporting definitions and of the main class types in a ROSS knowledge base:

[12] Baader, et al.: "The vocabulary consists of *concepts*, which denote sets of individuals …"

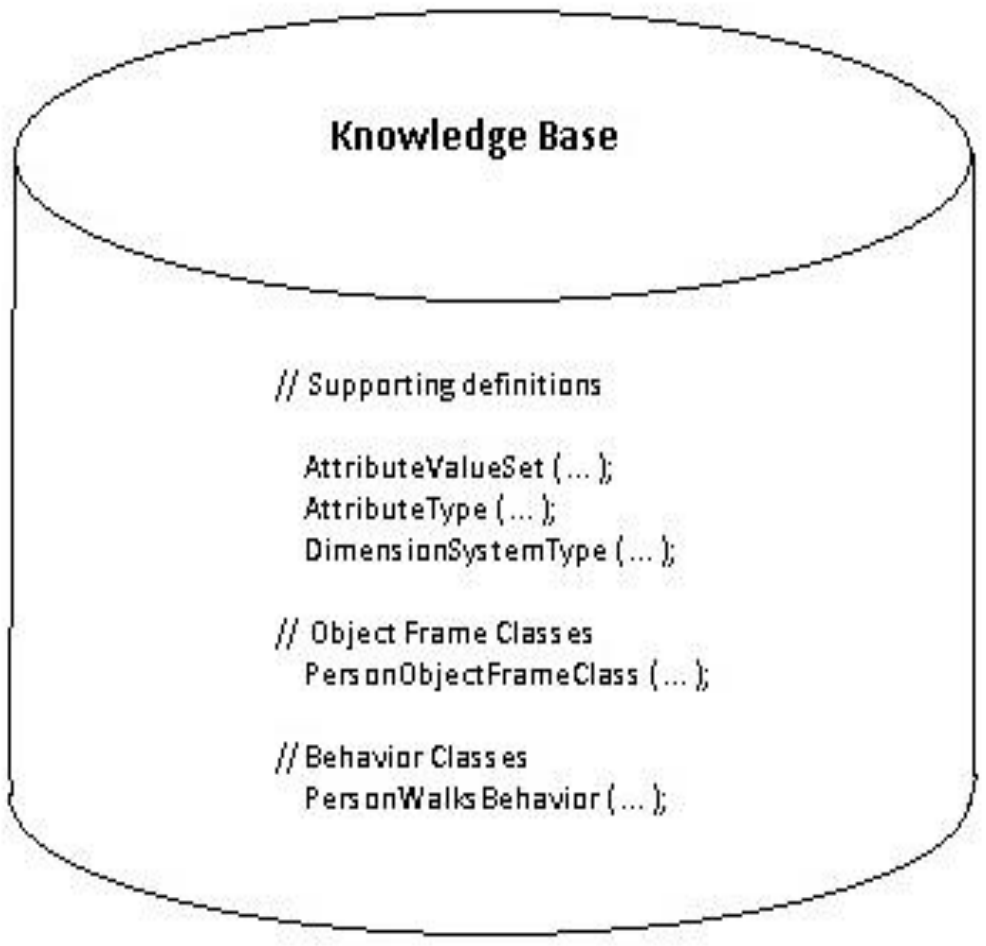

**Figure 2: Knowledge Base Definitions**

### 2.3 Inheritance

The most basic form of ROSS does not mandate any form of inheritance. The process of *instantiation* of an instance of a locational entity (either by a human using the ROSS approach or in an automated system) may involve the use of a class, but it is not necessary for this class to exist within a hierarchical inheritance structure. The minimal requirement for such a class that is used for instantiation of instances is that it contains dimension system information and a reference to a universal structural parent[13].

### 2.4 Grounding

The definitional classes of ROSS provide an important part of an answer to the *symbol grounding problem*[14]. In the field of natural language understanding, a related perspective was provided by Fillmore (1982). In his introduction to semantic frames, he states "By the term 'frame' I have in mind any system of concepts related in such a way that to understand any one of them you have to understand the whole structure in which it fits; when one of the things in such a structure is introduced into a text, or into a con-versation, all of the others are automatically made available". The definitions of ROSS provide this kind of infrastructure, so that when instantiation takes place, symbols have reference to more complex symbol-based constructs (the definitional classes).

[13] A structural parent class is an object frame class that is used for placement of other smaller embedded objects frame classes. The embedded object frame classes are usually not actual structural parts. The structural parent serves as a frame of reference.

[14] Harnad (1990) described the *symbol grounding problem* as the task of mapping symbols to aspects of the external world.

### 2.5 Granularity: "Grain Size"

ROSS is flexible with respect to the definitions that are used. Definitions are represented declaratively and thus an entire set of definitions is interchangeable with other similar such sets of definitions. The creation of a set of ROSS definitions starts with selection of one or more grain sizes (these are chosen per each class, or, where inheritance is used, the grain size can be defined for a higher-level base class). The grain sizes that are chosen for the base classes must be fine-grained enough to accommodate the information that is anticipated. For instance, a set of definitions that is used for news stories about events in the "everyday world" would require one set of grain sizes, whereas news stories about developments in the field of physics would require another set of grain sizes. (A single ROSS knowledge base can however accommodate all such needs if the higher level classes use appropriately-small grain sizes).

### 2.6 Primary Versus Secondary Information

ROSS maintains a distinction between "primary information" and "secondary information". Primary information is information that is specified in a way that is normalized with respect to a spatial/temporal frame of reference. It can be said to be "canonical"[15]. It is *entity-centric*, not *relationship-centric*. There are many types of secondary information, and ROSS subdivides these using a stratification scheme.

An example of secondary information is the relationship between two locational entities within a frame of reference. A specific example of a relationship within a transcript about past facts would be an expression that specifies that the spatial distance at some specified point in time between Person1 and Person2 is 3 feet. (In contrast, the primary information about Person1 consists of attributes that specify its location in relation to a structural parent class).

Primary (canonical form/normal form) information can exist within an object frame class or within a fact repository artifact.

### 2.7 Normal Form

For representation of factual information (excluding laws, rules, etc.), logic approaches can represent any assertion that can be represented in any other representational system.[16] However, when it is used to create fact artifacts, this expressive power can easily result in inconsistencies and redundancies. For example, a group of FOL expressions about a situation from a blocks world might mix together several distinct types of facts:

```
∃x:Block(x) – the predicate is a category

∃x:CompositionIsWood(x)
   - predicate is an attribute about the
     internal structure or composition

∃x,∃y:OnTopOf(x, y)
   - predicate is a relational attribute
```

In contrast the ROSS approach handles each of these cases in a different way. A ROSS artifact containing such facts would conform to normal form. In the examples shown here, for the first case, ROSS uses a class that has been fully and precisely defined in a definitions section of a knowledge base; in the second case, the assertion as expressed in ROSS would exist within a larger representational construct (primary information in normal form) that specifies not only the internal composition but also the location-related attributes of the object; in the third case the OnTopOf relationship would exist as part of a relationship construct that is handled as secondary information.

Normal form provides a unified way to represent information, with the result that ROSS knowledge bases and fact-based artifacts are non-redundant and complete – insofar as primary information is available. In some applications (e.g. NLU) where secondary information such as relationships is prevalent ROSS provides infrastructure that allows for storage of the secondary

---

[15] Woods (1975) (*III.B The Canonical Form Myth*) argues that canonical forms are not likely to be possible, (and are not necessarily desirable). The author's view is as follows: 1) for knowledge base definitional information: ROSS class representations are partly analogical; object frame classes do make use of primary information, in particular by the use of class attributes which can use shape templates, and 2) for fact repository artifacts: a canonical form is possible but it is constrained by the *availability* of primary information.

[16] For the sake of argument, neither the epistemological aspect (degree of belief or of certainty about the fact) nor the degree of truth (as in fuzzy logic) are considered here.

information in a way that is consistently integrated with the primary information.

### 2.8 Truth Theory Not Applicable to Primary Information

Negated assertions are absent from primary information: this has the implication that a ROSS class or fact repository instance does not participate in a truth theory or model. Brachman and Levesque (1985) discuss the spectrum of KR formalisms with respect to the level of expressiveness (from the simpler database schemes to the fully expressive FOL). The primary information form of ROSS is indeed closer to database, as it lacks universal quantification, negation, and disjunction. Primary information assertions are termed *simple assertions* that are true by default.

### 2.9 Simple Assertions

Simple assertions can be understood as *fact-like constructions*. By definition, a ROSS simple assertion specifies a *value* for a location entity. It may pertain to a class of things (e.g. as an attribute for a class), or it may apply to real past situations. Other uses of simple assertions are for the description of hypothetical facts in a hypothetical world. A distinction is made between simple assertions that represent completed states (or events) (whether real or hypothetical) and those that represent predictions[17] (predicted states) or goals (e.g. within the context of AI planning).

### 2.10 Stratification: Layers of Secondary Information

The first stratum of secondary information involves *disjunction*: it consists of disjunctive expressions and their variants (the variants include attribute value subsets and ranges). Examples include:

- An object frame class for “human parent” that contains a gender attribute involving values of male or female.
- A house cat class containing a structure section with an “animal head” part, where the spatial relational aspects are defined using a range of values specifying that the cat’s head is at located at least $\mathbf{d_1}$ inches but not more than $\mathbf{d_2}$ inches from the body.
- (in an NLU instance model) A house cat instance, where it is known from text input that is sitting on a mat: the precise location is not specified; it is somewhere (within a range) *on* and *above* the mat.

The second stratum involves *negation*. (Negation is viewed as a form of disjunction and therefore could be classified as part of the second stratum[18]). For example:

- The class of house cats that are ***not*** black – the attribute for fur color has a set of values that includes all possible colors except black.
- The house cat instance that is ***not*** on the mat: its actual location is a set of attribute values that range over *all other* locations in the represented world.

The following information types belong to the third stratum (alternately these are referred to as *tertiary* information).

- *Collections* (cf. universal quantification – the “for all x”)
- *Computed values*, for instance *counts* (e.g. the count of legs of a house cat instance).
- *Relationships*

### 2.11 Incomplete Information

Brachman and Levesque (1985) (citing earlier research) have tied the expressive power of FOL to its capability for dealing with knowledge that is *incomplete*. Disjunctive and negated expressions are exemplary of such information. The objective of ROSS is not to subvert fundamental epistemic tenets; rather it involves the full *exploitation* of the information that is available in a situation or domain. This exploitation of primary information is best illustrated by the use of object frame classes with their capability for rich specificity.

### 2.12 Axioms and Rules

The question of whether axioms in a KR scheme or a logic are actually *rules*, or something more fundamental and intrinsic, is not addressed here. It suffices to state that, where relevant, ROSS makes differentiations between the following categories of axiomatic, or “rule-like” concepts and constructs:

- *Definitional axioms* (including set-theory axioms) (e.g. transitivity of *PartOf*:

[17] “Prediction” here refers to a predicted state, a sort of “future fact”. This definition differs from the use of “prediction” in the context of machine learning.

[18] This view is possible based on the use of attribute value sets that are finite sets.

"$\text{PartOf}(\alpha,\beta)\ \Lambda\ \text{PartOf}(\beta,\gamma) \rightarrow \text{PartOf}(\alpha,\gamma)$")

- *Dimension system axioms/postulates* (cf. geometrical axioms, e.g. the Pythagorean theorem)
- *Computational axioms/postulates* (example include the axiom "2 + 2 = 4", and axioms about prime numbers)
- *Correlative rules*: these includes rules that model causal phenomena (probabilistically or otherwise) and rules that model statistical correlations.

### 2.13 Correlative Rules

A ROSS correlative rule is a representational construct that represents correlation in a problem domain. Correlation may or may not involve causality (i.e. the *laws* for some domain). There is not a limit on the types of correlative rules that can be constructed using the ROSS KR scheme as a foundation - this is due to the view that inference (reasoning) is a multifaceted set of tasks that should not be overly constrained by predefined approaches. Rules are not a part of ROSS fact repository artifacts: since rules are handled separately from facts and other fact-like constructs, a variety of *rule base* approaches are possible.

### 2.14 The ROSS Perspective on Implication and Entailment

The tradition within the AI KR&R field has usually involved a model of *entailment* that involves a single knowledge base ("KB") that contains a mixture of definitions/classes, facts and rules. ROSS handles this somewhat differently as is explained here.

Where a ROSS class or instance is limited - involving a simple assertion (primary information in normal form), or a disjunction of simple assertions, and the negation operator is not present - the concept of implication is limited to a form of meta-information reasoning that uses definitional axioms, or dimension-system axioms. (This is a sort of "static" reasoning that can be performed off-line if necessary in a given implementation).

Where the negation stratum is involved, the law for implication $((\alpha \rightarrow \beta) \leftrightarrow (\neg\alpha \vee \beta))$ is relevant, and a ROSS processor can make use of this in order to perform inference.

Although ROSS sharply differentiates definitions (ROSS KB) from facts (fact repository artifacts), the above reasoning can involve either a ROSS KB or fact artifact.

### 2.15 Relating Antecedent to Consequent in a Correlative Rule

In the field of logic, *connexive logic* (e.g. *relevance logic)* addresses the need for correlating information in the antecedent of a rule with information in the consequent. ROSS formalizes the concept of associating antecedent with consequent using a representational construct that is a part of all correlative rules, referred to as the *binder*. A binder is an abstraction that is implemented in such a way that the locational attributes of entities in the antecedent of a rule are related to the locational attributes of entities in the consequent of the same rule. Otherwise ROSS does not place unnecessary restrictions on the structure of correlative rules; the resultant sophisticated rule structures are viewed as the basis for automated reasoning that has a corresponding level of complexity and sophistication.

### 2.16 Instantiation

Within the context of automation, instantiation is the process of creating an instance of a class within some fact-containing repository such as an NLU instance model. The instance is based on the class.

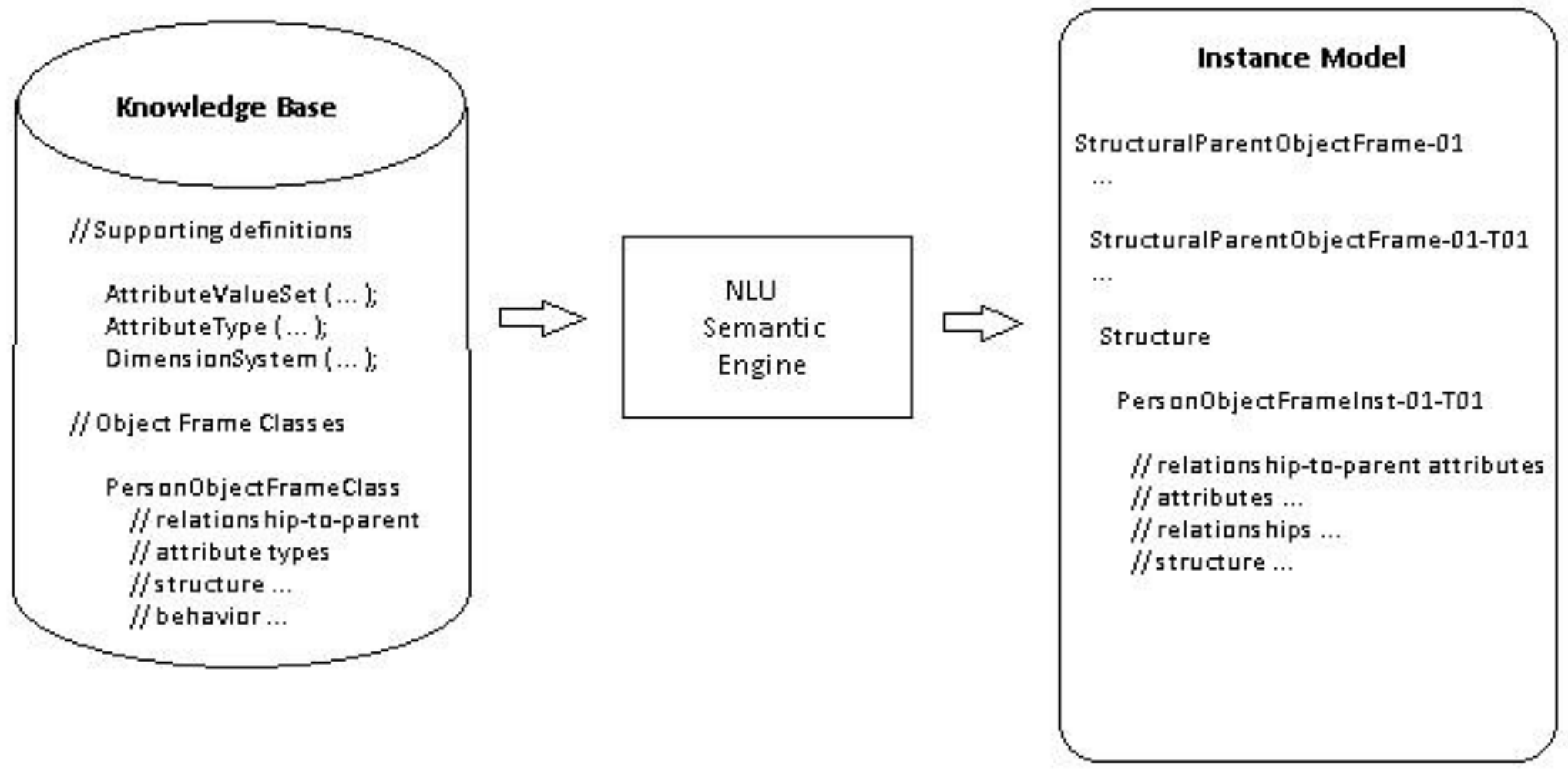

**Figure 3: Instantiation in ROSS**

**Figure 3** illustrates the general concept of instantiation by showing how it takes place within an NLU system. (In an NLU system, the task of instantiation is performed by a "semantic engine").

### 2.17 Organizational Advantages of a Physical Symbol-based Approach

ROSS is the basis for a symbolic AI approach for both knowledge base abstract definitions and for fact transcripts and instance models. Knowledge base definitions are written in the Star language. Instance models are XML documents. A symbol-based approach has value in that it is very effective as a way of organizing information, and at the same time it is flexible in allowing references to or inclusion of non-symbolic representations. For instance, use of the XML standard for instance models allows for the specification of uniform resource identifiers (URIs) that specify abstract resources that are external to the XML document. (Example resources would include binary data objects, e.g. a file containing a bitmap, and web-based services).

### 2.18 Representation of Human Agents

A premise of ROSS is that aspects of human agency can be represented using the same underlying approach that is used for representing any other situation in the physical world. There are no special "meta-representational" constructs in ROSS that would handle the domain of human agent cognitive processes differently from anything else in the physical world. The subjects of this type of representation include representational entities and processes themselves – wherever the problem situation or natural language subject matter consists of human thoughts (e.g. plans or intentions), and they include instances or artifacts of human communication (e.g. spoken sentences). Mental computational processes that are performed by humans are represented in a similar way; automated computational processes are also represented. Representations in ROSS of human or automated representations or computational processes are typically ones that utilize more-abstract physical spatial coordinates or spatial location dimensions. For example, to represent the fact that an intelligent agent has a cognitive representational concept that represents "vehicles", it may be sufficient to describe the physical location of this concept as existing in the agent's memory and that it is spatially distinct from other related concepts. (Enumerated values that express spatial locations are useful for this type of situation).

### 2.19 Application to Abstract Areas

ROSS can handle representations of abstractions involving entities that have physical attributes that are not relevant to a given domain. A fuller treatment is given in the section that describes the features of ROSS.

### 2.20 Star Language

ROSS includes a language referred to as the Star language ("Star" is derived from the word "structure"). The Star language is used as the means for encoding the definitions, and as an alternate form (versus XML) for encoding instance models. The Star language was originally developed by the author as a computer software specification language; its role within ROSS has been expanded to allow representation at the widest level, i.e. covering all problem domains or domains of discourse.

### 2.21 Knowledge Bases

A ROSS knowledge base contains class definitions, behavior definitions and other supporting definitions that use the Star language. For convenience the term "Infopedia" is used to refer to such a knowledge base. (A knowledge base that is used for automated inference also contains rules). Examples classes from the NLU area include an object frame class called "PersonClass", and a behavior class called "PersonHitsPerson".

### 2.22 Fact Transcripts and Instance Models

A ROSS *fact repository* is a representational artifact that contains fact-like constructs. The following are two important categories of fact repositories:

- a *specification transcript*, or *transcript* contains a collection of related fact-like constructs that exist for some storage or computational purpose. For instance it may be used in the context of automated reasoning (an example from AI planning is the specification transcript that contains specifications of predicted states, conditions and goals)
- *instance models* contain fact-like constructs and that are used in the area of NLU for meaning representation.

In the area of AI automated reasoning, various transcript types have been developed by the author: these include goal statement transcripts for computer software requirements and design specifications (this transcript looks somewhat similar to a computer program), and transcripts about past facts for a diagnostic expert system.

### 2.23 Foundations for Correlative Inference

Correlative inference rules involve correlations between attribute values of at least two distinct locational entities (involving the binder concept). This type of reasoning include the following:

- Reasoning that uses rules that represent causality for a problem domain. Such rules may be "if/then" rules that involve descriptions of causes and effects that are propositional, or the rules may express a functional relationship.
- Reasoning that uses probabilistic rules that are descriptive of correlations that have been derived about a problem domain.

### 2.24 Applications Within Implemented Systems

ROSS has been successfully applied in three separate systems: an expert system computer program generator (Hofford, 2001 and 2010), an expert system for diagnosis of network faults (Hofford, 2013), and a representational technique that exists in a natural language understanding system (referred to as "ModelBuilder"). The ModelBuilder NLU system uses ROSS Star language definitions and generates instance models that use XML as an encoding format.

## 3 Ontology

The ontological commitments of ROSS involve several conventions; these include a set of constraints and an important requirement. These modeling restrictions and requirements are essential for producing knowledge base and fact artifact representations that are structured. The ontology is restrictive for a purpose: traditional upper ontologies that involve a taxonomic tree of objects with "thing", "substance", (or some variation thereof) at the root are not viewed as relevant for the purpose of organizing the *represented world* as it is addressed by ROSS.

The study of the "categories" has a long tradition in philosophy – on that predates Aristotle. Aristotle described an upper ontology in his *Categories*. Tree-like ontologies have been predominant ever since and are still prevalent.

Russell (1945) was a critic of Aristotle's taxonomy; he provides an interesting perspective:

> *"I do not myself believe that the term "category" is in any way useful in philosophy, as representing any clear idea. There are, in Aristotle, ten categories: substance, quantity, quality, relation, place, time, position, state, action, and affectation… There is no suggestion of any principle on which the list of ten categories has been compiled. …*
>
> *I conclude that the Aristotelian doctrines with which we have been concerned in this chapter are*

*wholly false, with the exception of the formal theory of the syllogism, which is unimportant".*

Regarding the basis for a set of categories, the question of "what exists" has been addressed throughout the history of philosophy. The influential work of Quine (1948) acknowledges two kinds of existence: physical things, like the continent of Australia, and abstract things – e.g. prime numbers.

In contrast to the approaches that have been described, the ROSS approach is not based on a tree-like taxonomy at all. If a taxonomy of "things" ("what exists") were drawn, ROSS would only have one entity at the root, the unit-sized location entity.

Because the ROSS ontology is itself a representational *tool*, many philosophical questions simply are not addressed – it is intended as an ontological framework that has been integrated with a representational methodology. The ROSS ontology is a sort of "pseudo-ontology": it is the basis for a representational framework that is useful for a broad range of domains.

Adherence to the ROSS ontological conventions facilitates modeling practices that ultimately result in definitions and expressions that are internally consistent, unambiguous, and non-redundant. These conventions apply to the modeling of the universe of discourse or a domain of discourse or problem domain. The ontological constraints are as follows.

### 3.1 Fundamental Ontological Constraint

The first and most fundamental ontological convention is a constraint: a unit-sized location entity is the most basic – and only - thing that exists in a represented world. It is a single-time location within a dimension system (it is *transitory*). Such unit-sized entities do not move - the intuitive concept of *movable object* is not a first-class object of this ontology. Movable objects must be represented as an aggregation of unit-size objects. Likewise, anything else – entities, motion, state changes, etc. - that is represented using ROSS must build on the fundamental building block of the unit-sized location entity.[19]

An implication of this constraint for ROSS representation is that unit-sized location entities are the only thing that can be existentially quantified. (As will be explained however, within fact-containing repositories, ROSS does not actually use existential quantification).

### 3.2 Second Ontological Constraint

The second convention is also a constraint: a unit-sized location entity does not contain an "object"; rather, the *value* of a location object is represented as consisting of a numerical value from the set of natural numbers. The Star language provides the infrastructure for describing sets of values using the *attribute value set* statement. As will be seen, other representations can be used in place of integers, such as enumerated values or string values.

An example of a fixed location with a specific value at a specific point in time involves the digits of the national debt clock (any particular digit) in New York City. The concept of a fixed location with a specific value at a specific point in time in ROSS can be seen as a generalization of this and many other instances. However as a generalization it extends even to movable objects such as a falling apple or a bouncing ball.

### 3.3 An Ontological Requirement

The third convention is a requirement, and is related to the first constraint of the ontology: the universe or domain of discourse is viewed as consisting of at least one dimension system, so that at the least, a unit-sized object, designated "Object-A", has a location that is distinct from the location of one, some or all other objects. This requirement results in representations that contain infrastructure that allows all entities to be correlated with each other within a central frame of reference. A dimension system can be very abstract, as is the case where specifics of a location do not need to be represented. Or, a dimension system can be very specific, for instance, it can be the implementation of a four-dimensional coordinate system with x, y, z, and t coordinates.

Why is the dimension system required for ROSS? The requirement for the existence of a dimension system is the foundation for an entire set of representational constructs that represent structure. A dimension system provides a frame of reference for all entities that are described.[20]

---

[19] The unit-size location entity does have a single unit of duration in time. The term "transitory" refers not to an instantaneous point in time, but to an interval of time.

[20] Note that ROSS also contains the *relationship* construct (as secondary information) that is useful for representing specific relationships between two entities. An example would involve a person and a car in a parking lot: each of the objects (person and car) may be directly related to the object frame class for the

### 3.4 Implicit Existential Quantification

An implication of the above ontological requirement for dimension systems is as follows. The act of creating a structural parent object frame instance within a fact-containing repository (e.g. an NLU instance model) results in the existence of a dimension system that is housed within the structural parent instance (e.g. a Cartesian coordinate system). This can be visualized as a rectangular shaped region (a cube or rectangular right prism (a cuboid)). It is a collection, or aggregation of unit-sized location entity instances. Because this frame of reference exists, there is not a need for the use of the existential quantifier - as would be the case with FOL – for propositional ROSS expressions. Once this frame of reference has been created, the main subsequent representational task is that of infusing or populating the individual cells (like the cells in a matrix) with values.

### 3.5 Atomism

The earliest known proponents of an atomic theory appear to have been the Greek philosophers Leucippus and Democritus (Guthrie, 1950). Their concept of atoms included the idea of indivisibility. (It is noteworthy that atomism was rejected by Aristotle and that the concept was dormant until the 19th century[21]). The Greek word *atomai* means "unsplittable" and was used to describe the smallest existent particles - these particles were hypothesized as being solid, hard and indestructible. The modern physics concept of atom is more specific due to better detection devices (not to mention the body of analytic and empirical methods of modern science) – an atom is precisely defined with respect to its attributes, such as size and mass.

The ROSS approach borrows one aspect from Greek atomism: the unit-sized location entity of ROSS is *indivisible*. Indivisibility of ROSS atoms is necessary so that value set values can be used as descriptors. If a ROSS unit-sized location entity were divisible, a new set of attribute value sets would need to be created – one for each of the newly-created components.

A key difference is that the ROSS "atom" comes into existence, last for only one time interval, and cannot be said to exist again. It has – at the least - a numeric value. Insofar as the value can vary from one "atom" to the next, the ROSS concept in this respect also differs from that of the early Greeks and from the physics atom, since both concepts involve a uniformity of all atoms.

The comparison between ROSS unit-sized entities and atoms from the field of physics breaks down with respect to specificity. Since ROSS is only a representational scheme, it does not involve any implicit claim that a unit-sized location entity actually exists. This allows for complete flexibility in modeling, as an atomic size may be chosen that is capable of representation for a problem domain or for a wide range of problem domains.

### 3.6 Root size spatial unit is a black box

Tarski (1927) described a theory wherein a primitive - the smallest indivisible "thing" that exists - is a sphere. ROSS makes the ontological assumption that the shape of the primitive (smallest) existent entity for a given domain does not actually matter. For purposes of creating representations, it is only important to differentiate the *value* of the smallest *locational* entity from the values of other locational entities. In essence then, the internal composition of the smallest spatial unit (whatever this is chosen to be) is unimportant – it is a "black box". (However, it should be noted that when a Cartesian coordinate system is used, the unavoidable conclusion is that the shape is that of a *cube*).

### 3.7 The Treatment of "Space"

As described above, a transitory unit-sized location entity can only be described in terms of its *locational* properties and in terms of its internal *value* properties. A consequence of this is that "space" is not a special case. Space is treated as a value, no differently from other values. Whether or not a unit-sized location entity has a value of "space" is important in several areas: for "shape patterns", and for rules for inference. (To accommodate these uses the Star language has a set of built-in value categories called "Space" and "NotSpace").

### 3.8 Substance

Substance is not given any special handling: substance is modeled in a way similar to motion – it is described in terms of attribute "*value*" values.

---

larger parking lot using structural relationship-to-parent attributes, but at the same time their explicit spatial relationship to each other may be represented in order to capture the meaning of the phrase "the woman stood near the car".

[21] Cf. Sowa (2000) - Aristotle's concept of *monads* (units), specifically his idea of point and instant are somewhat similar to atoms.

From the ROSS perspective, substance does not necessarily "exist" at all. Rather, substance as a representational concept is a result of the modeling of location entities and their values, and of the causal phenomena that are associated with them.

### 3.9 Time as Interval with Duration

Ontologies and representations of time are a much-researched area and detailed background is beyond the scope of this paper[22].

ROSS is based on a particular view wherein time consists of a sequence of intervals during which motion does not take place, in other words, objects are "frozen". Subsequent to each interval, and instantaneously, objects then "jump" to the next location (or they remain in the same spatial location if they are not moving). This is not an ontological presupposition – it is the tenet for a representational approach. The question of whether or not the physical world actually behaves this way is left to the field of physics.

### 3.10 Motion and Process

Motion is viewed as an aggregation of unit-sized location entities. The following example is pseudo-code that illustrates this concept, where t1 and t2 are the labels for two intervals along a timeline, and x, y and z values are integer-based coordinate values (note that the two location entities are adjacent). This represents some *thing* that "moves" from x1 to x2.

- Entity-1-t1 at x1,y1,z1 @ t1 // value = solid
- Entity-2-t1 at x2,y1,z1 @ t1 // value = space
- Entity-1-t2 at x1,y1,z1 @ t2 // value = space
- Entity-2-t2 at x2,y1,z1 @ t2 // value = solid

### 3.11 Digital View

The ROSS approach can also be understood by way of comparison with approaches that involve a digital view of some domain. For instance, acoustic digital sampling produces data that represents a sound pattern at discrete time intervals. ROSS is similar in how it represents real world objects and events: it abstracts from the problem domain in order to create representational constructs that have application to discrete space and time intervals.

[22] Some references for this area include Allen (1983), Hayes (1996), Shanahan (1997), Sowa (2000), Russell and Norvig (2003).

### 3.12 Representation of Shape and "Telescopic" Dimension Systems

Shape is not a primitive concept in ROSS. However shape is addressed as part of two areas:

- Classes can be defined in terms of shape. For instance, a class that represents containers, e.g. cups, that holds liquids must be defined with respect to its shape.
- Rules that use ROSS as their underlying representation can involve specifications of shape.

The use of shape within a class definition is as follows:

- A dimension system and a specification system that uses it are defined. (This discussion will use a dimension system that is a 3D Cartesian coordinate system). The specification system inner content section must incorporate a qualitative attribute type that uses an attribute value set that contains at least one value that inherits from the Space value category and at least one value that inherits from the NotSpace value category.
- An aggregate object frame class with a type of range is defined. (A range object frame class is not defined in terms of named structural sub-parts but rather it is a group of spatially adjacent unit-sized locations, along a line, in a plane, or within a 3D cubical solid).
- *Smallest possible shape*: A pattern is created that represents a shape using as few locations as possible. (Note that in ROSS, a pattern can be minimally defined as one that involves a single unit-sized location entity with a specific value – an example is a single pixel on a computer screen with a color of red). A shape-representing pattern (ShapePattern) involves an aggregation of spatially-adjacent unit-sized locations each of which have a specific value that is only distinguished from other values with respect to whether or not it is a member of the Space or NotSpace value category.
- Note that a ROSS *template* is an aggregation of values or a mechanism for generating such values (e.g. a drawing instruction set) that involves the possibility of disparate values. For example a template for a house cat would likely use a variety of colors or compositional attributes. In contrast, a ROSS ShapePattern is limited to containing only values from the Space and NonSpace categories.

- *Use of ShapePattern in class definitions, and magnification*: a ROSS ShapePattern is used within a range object frame class, within the Attributes section, in order to specify the shape of the class in an abstract way that is not tied to specific spatial sizes. The process of instantiation uses the class and its shape pattern to create an instance: this process must specify specific spatial dimensions and therefore it involves a computational task called *magnification,* using the shape pattern in order to infuse the instance with a set of NotSpace values that correspond to the original NotSpace values of the ShapePattern.
- A telescopic dimension system mapping is used in order to accomplish the magnification of a shape pattern to generate an instance pattern for the shape.

The ShapePattern as described above can also be used in rules.

### 3.13 The Diorama Analogy

A diorama is a small-scale replica of a scene. The diorama presents a useful analogy for the structure that is instantiated when a ROSS NLU instance model (or KRR fact transcript) is created. The diorama typically has a physical frame – a ROSS *instance model* has a single *structural parent object frame class instance*, which provides a framework, and more specifically a dimension system within which to "put" things. The difference involves the aspect of time: an NLU instance model *context*[23] contains a timeline, and changes can take place, whereas the diorama is a way in which to present a static visual scene.

#### 3.13.1 The Cartoon Analogy

The modern cartoon provides the second analogy for ROSS representation instances. Cartoons achieve the appearance of motion through the utilization of fixed-location images. The continuous aspect of motion thus is not actually involved. ROSS as a representation technique uses the same fundamental approach.

### 3.14 Convenience Assumptions

ROSS includes several concepts that can be used in the representation of facts – these are assumptions the use of which precludes the necessity for the excessive use of explicit assertions.

[23] A context contains a top-level structural parent. (Cf. the section on Instance Models for a full definition of the *context).*

### 3.15 The Empty Space Assumption

The *empty space assumption* has relevance specifically for fact repositories. This is the assumption that an instantiated instance has a value (if it is a unit-sized object instance) or consists of values that are members of the Space value category, unless specifically specified otherwise as is done with infusion. (By way of analogy, this is like placing a wire frame model into a diorama – it is initially empty, until it gets filled with something).

### 3.16 Perpetuation

The perpetuation assumption involves perpetuation of values along a time line; it can be used in similar fashion as the empty space assumption: the assumption is that for any unit-sized location that has been infused with a value at time t=n, it can be assumed that the subsequent unit-sized location at the same spatial location (at time t=(n+1)) will have the same value unless it is overtly specified to have a different value. This assumption is useful for stationary objects but does not address the representation of objects in motion.

### 3.17 The Frame Problem

The frame problem is relevant for systems that perform inference; it is addressed in ROSS by the above two assumptions. At a deeper level, the frame problem does not present the same challenge to ROSS that it does for FOL-based systems due to the existence (in ROSS fact repositories) of a structural parent instance: by implication this involves the existence of all of its unit-sized location entities (cells). Global assumptions like the empty space assumption and the perpetuation assumption are used in order to conveniently specify most values. Values that change through time (as part of state changes or motion) are explicitly set as indicated by the inference process.

### 3.18 Dimension System Concepts

ROSS uses the term "dimension system", rather than "coordinate system" as a way of denoting that a dimension system is a flexible concept: it need not be tied to any pre-defined method for the representation of space and time (e.g. a Cartesian coordinate system). The basis for the dimension system concept is a cognitive process referred to here as *segmentation.* Segmentation is defined as the cognitive process of differentiating locations from each other. An instance of a pro-cess of binary segmentation yields two location entities: each entity can be represented with a symbol but little else is needed; for instance there is no "directionality" in the absence of a third reference point. However multi-part segmentation has the following implications: 1) general directionality emerges as soon as three entities exist, 2) numeric representation of locations using natural numbers (or integers) is useful, 3) an origin is needed for each dimension, and 4) specific directionality (within a dimension) (due to the existence of an origin and a numeric scheme) implies a set of relationships (e.g. "less than" and "greater than").

Dimension systems can be mapped from one to another, i.e. a set of specifiers that use one dimension system can be transformed into an equivalent set of specifiers that use another dimension system (e.g. geographical coordinates that use latitude and longitude can be transformed into Cartesian coordinates).

### 3.19 What are the ROSS Primitives?

The discussion of the ROSS ontological foundation is not complete without addressing the topic of *primitives*. From one perspective there is only one primitive in ROSS: the unit-sized location object; all other primitives are built as aggregations of this primitive. Nevertheless it is conceptually useful to delineate some of the aggregations that can be built and to relate other primitives that are abstractions to ROSS.

There are two categories of ROSS aggregation primitives: static primitives and dynamic primitives. The static primitive classes are as follows:

- The unit sized location entity as discussed above
- The range object frame class (it can be understood as a 3D array)
- The aggregate object frame class (containing named structural components/parts)

The dynamic primitive classes are:

- The state change class
- The behavior class
- The 4D structural parent *instance* (containing multiple time points)

What may appear to be missing are the following:

- Commonsense objects that move
- Motion and processes
- Events
- Agent actions
- Cause, effect
- Potential, force

Commonsense objects and motion have been described (process is viewed as equivalent to motion). An event is a state change; an agent action is a process. Cause and effect are treated as dynamic aspects and are handled using rules. Potential and force are viewed as abstractions that augment the conceptualization of causality.

### 3.20 Comparisons

The work of Hayes (1985) in the area of qualitative physics uses a four-dimensional approach. He states the following:

> *A physical object is a three-dimensional entity which has an associated history representing the life-span of the object: a slice of this history (which we will call the* **life** *of the object),* **is** *the object at a given time.*

Hayes' model of the physical world (for the purpose of describing liquids) is close to that of ROSS. A key distinction is that Hayes appears to preserve the concept of objects that exist through time whereas ROSS does not permit this.

Johnston (2011) provides an overview of the objectives and architecture of the *Comirit Objects* project, a system that represents and reasons about things in the physical world. The knowledge representation approach centers around the *voxel*, which is defined as "the 3-dimensional equivalent of a pixel: it refers to a small cubic region in 3D space." Further definition is provided "Voxels are the 3D equivalent of bit-mapped or raster images." The use of voxels for representation of physical objects is as follows "Instead of representing an object by a complex polygonal structure, its shape can be approximated by a set of voxels that fill a similar space." The voxel concept shares a basic similarity with the ROSS unit-sized location entity. Whereas the characteristic of color can be attributed to a voxel, ROSS's implementation of the unit-sized location entity – the object frame class – is capable of association with any attribute that can be conceptualized (e.g. material composition, texture, mass). Further detailed comparison is beyond the scope of this paper: it should be noted that voxels can be described with meta-data that specifies structure and shape; this may correspond to the features of ROSS for handling structure and shape.

Regarding representations of the dynamic or behavioral aspects, Johnston states "Voxels will not be able to describe the workings of intricate machinery but they can capture the approximate shape of such machinery. If necessary, dynamics of machinery can be captured as a collection of models that show the sequence of actions." The ROSS approach parallels this capability with the *behavior class*.

### 3.21 Summary of Ontology

The ontology of ROSS is rigid for a purpose: consistent adherence to these conventions and the afore-mentioned requirement serve as guidelines for modeling as it relates to creating definitions. The knowledge engineering task of creating definitions can be performed by a human or it can be automated (automated approaches are briefly described in the later section on knowledge acquisition).

## 4 Summary of ROSS Features

The ROSS features include the features of the Star language that are used for definitions, and features that are relevant for fact repository artifacts, e.g. instance models. The Star language features borrow to some extent from object-oriented programming languages such as C++.

### 4.1 Specification with Natural Numbers and Integers

Numeric attribute values are either natural numbers or integers. Numeric attribute value sets that are used for locational attribute types must be members of the set of integers. Numeric attribute value sets that are used for qualitative attribute types must be finite subsets of the set of natural numbers. Within the primary information section of a fact repository artifact, where data or natural language text represents real numbers, or where values are computed (e.g. by division) to yield a real number, rounding or truncation of numeric values must take place. (Note: the use of real numbers as secondary information is not precluded from the ROSS model).

### 4.2 Feature: Attribute Value Ranges

Attribute value ranges have an important use within definitions of object frame classes – they allow for a component to be located approximately within the parent structure. For instance, a class called "FrontEngineAutomobile" would specify that the engine compartment is situated within a certain section of the parent class (the Automobile). The class definition does not specify the exact location – it is specified using a range of values (e.g. within the range of 5 cm to 100 cm from the front end of the car). Instances that are based on the class can specify the exact location if this information is relevant.

### 4.3 Feature: Star Language Definitions

Star language definitional constructs allow for the definition of attribute types, attribute value sets, and other definitional representational constructs that are used by a human or by an automated engine to construct meaning representations. An example of the Star language definitions is the *dimension system*, a unique feature of ROSS: it is a set of integrated definitional constructs that provides one or multiple *locational* attribute type definitions that must be used together in expressions that specify the location of a location object (i.e. an object frame class instance).

### 4.4 Feature: Inheritance

A distinction is made between entity classes that are used to directly instantiate instance entities, and higher level entity classes in an inheritance hierarchy. As has been stated, higher level classes and inheritance are important but optional features of ROSS and the Star language. However, the instantiation of an instance does require the use of at least one class. For use in generating an instance model, the typical case involves using an object frame class (an entity class), such as *CarObjectFrameClass*, which class either may contain all information, or it may rely on higher classes to contribute some of its features. The alternate case involves using a more-abstract object frame class referred to as *ObjectObjectFrameClass*. This class is abstract with the exception of dimension system information and a reference to a universal structural parent. For example, an NLU instance model may contain an instance that has been instantiated based on *ObjectObjectFrameClass*. Since the structural parent is known it can be inserted into the model. Subsequent contextual information in the input may subsequently be used to add attributive or behavioral information about the object instance.

### 4.5 Feature: Dimension System

The next feature group is based on the ontological requirement that at least one dimension system must be chosen. The Star language feature is referred to as the *DimensionSystem* and it consists of a set of related location attribute types (e.g. x-coordinate, y-coordinate, z-coordinate, time). The location attributes describe where an entity is in space and time. When these defini-tions are used in generating or creating an instance model, type checking can and should be performed to ensure that each attribute type is used (in an instance model or fact transcript construct called the "*dimension set expression*") – this provides for representation expressions that conform to the ROSS requirements for the specification of structure.

### 4.6 Feature: Mapping Between Dimension Systems

Star includes a concept that allows for mapping between dimension systems. For instance this allows *common-use everyday* relational attributes to be used where greater precision is not needed – e.g. "Arizona is in the south west region of the United States". The enumerated value that represents "south west region" can be mapped to a more precise set of locational coordinates such as latitude and longitude. This feature is especially important for natural language story understanding because human dialog is often imprecise with respect to place and time.

The ability to translate or map between different dimension systems is viewed as a key feature of ROSS; it is a reflection of how human memory and cognition work. Humans seem to have a general-purpose three-dimensional frame of reference that underlies (perhaps all) representations. The capability for representing the location of things in the physical world often involves smaller "customized" frames of reference. An example would be a mental representation that a particular house is at 1000 State Street, in some city. The custom frame of reference has a dimension system that consists of city identification (name), street name and street number. There is a mental capacity for going back and forth between this custom representation and the master (3D) frame of reference.

### 4.7 Feature: Specification System

The *specification system* definitional construct contains definitions in each of two sections: the *dimension system* section and the *inner content* section. The dimension system has already been described. The inner content section either defines a set of attribute types that describes the value of an entity (e.g. the car is blue), or it is a specification of component-wise structure. The *specification set expression* uses a specification system similar to how the dimension set expression uses a dimension system.

### 4.8 Feature: Aggregate Representational Constructs

The Star language *ObjectFrameClass* construct allows for the representation of a spatially adjacent aggregation of unit-size objects. An object frame class may represent a single unit-sized object frame or it may represent an aggregation of such units in which case it has the shape of a 3D cuboid.

### 4.9 Feature: Structure

There are two ways to represent mereological structure in ROSS. First, a group of representational constructs enable the representation of *structure* that involves *components*: these include the structural parent entity, a set of "relationship to parent" locational attributes and a structure section for aggregate entities that models compositional ("PartOf") attributes. The infrastructure for representing this type of structure is complex; these features are explained in detail in the upcoming section on Star language definitions. The second approach is the object frame class range, described next.

### 4.10 Feature: Object Frame Class Range

The object frame class range is a special type of aggregate object frame class (composed of multiple spatially adjacent unit-sized location entities that span one, two or all three dimensions). This construct does not have in internal structure that is composed of structural components; rather, it is a sort of drawing canvas on which a picture can be drawn. A simple example would involve a cubical object frame class range in which can be drawn a sphere. The representational construct that is used for the drawing is called a *template* class (described in the upcoming section on Star language features).

### 4.11 Feature: Multiple Parallel Structures to Support Drill-Down Analysis

The human cognitive capability for analyzing the structure of a physical object in a "drill-down" fashion is the basis for a ROSS feature that allows for multiple representations of the same spatial structural area (cuboid region). An example involves a part of a push lawn mower: a mower blade is modeled as a structural component that contains individual parts such as the blade edge; however a parallel structure section consists of a single object frame range class (e.g. using a millimeter grain size). The object frame class range is modeled using a ROSS template, which in turn is infused using either a set of drawing instructions or a 3D bitmap.

### 4.12 Feature: Higher Classes

The *higher class* construct allows for the creation of hierarchies of classes. Class inheritance is viewed solely as a way of aggregating or consolidating groups of attributes and structural features – it is a convenience mechanism, nothing more - higher classes (parent classes) supply additional representational information about a given class. ROSS allows for multiple inheritance (multiple parent classes per class).

ROSS also allows for multiple similar, or parallel, classes, based on the view that some classes, such as a "person class" are not necessarily a single class, but are so potentially complex that the use of multiple such classes might be needed in order to model a variety of feature collections. Different domains would use different such classes. An example might involve several classes such as *PersonAsCountryCitizenClass*, *PersonAsBiologicalLivingEntityClass*, and *PersonAsTravellerClass*. Each of these would be useful in different domains.

### 4.13 Feature: The Attribute

The Star *attribute* is an expression that closely corresponds to a FOL atomic sentence that contains a predicate and a term that is a single constant. An attribute is a strongly typed two-part construct. It consists of a pre-defined attribute type name and an attribute value that is a member of a pre-defined attribute value set. The ROSS notion of attribute type and attribute value is roughly the same as that which has been in widespread use in software applications for many years, for instance, the attribute from the field of logical data modeling for databases.

An example attribute is:

```
<Attribute ref = VehicleExteriorColor
           val = "Silver" />
```

The attribute is composed of the "Attribute" keyword, then the "ref" keyword ("reference") followed by an equal sign and a defined attribute type name, and the "val" keyword ("value") followed by a value that had been defined as a member of an attribute value set (the attribute value set that was defined within an *attribute type* called "VehicleExteriorColor").

The use of attributes rather than predicates (as with logic) provides for a set of criteria for indexing. For instance, the logic assertion "E(x): Blue(x)" (which can be read "there exists an object such that the object is blue"), makes use of a predicate that actually corresponds to a Star attribute *value*, not the attribute type name. Attribute values do not provide a good basis for indexability since they may be members of very large sets. In contrast attribute types are more appropriate as criteria for indexing as it is needed to support queries of structured information.

ROSS attributes are not limited to containing constant values: an attribute value may consist of a value *range*, a math expression, or a function name that refers to a function that has been defined within the ROSS knowledge base. (An attribute value may also be a reference to a defined template).

### 4.14 Feature: The Two-part Attribute Cluster

The ROSS *two-part attribute cluster* is a conceptual feature that can take any of a variety of forms. The two-part attribute cluster satisfies the intuitive concept of a *fully specified fact*: it represents both the *location* and the *value* of an entity that exists in a 4D represented world. A two-part attribute cluster can exist within a class definition in a knowledge base or it may exist within a fact repository artifact. The Star language implementation of the two-part attribute cluster is a representational construct that consists of at least one attribute from the attribute super-type called "location attribute types", and at least one attribute from the attribute super-type called "value attribute types".[24] The rationale behind this requirement is that it produces ROSS expressions that *fully* describe entities from the represented world. (The XML implementation of a two-part attribute cluster, e.g. in an NLU instance model, is not shown here).

The two-part attribute cluster is the equivalent of a set of propositions or assertions in logic; where these assertions would include one or more propositions that represent the *location* of an entity and one or more propositions that represent the *value* of the same entity.

The following is an example of a two-part attribute cluster. (The Star language fragment also shows several preliminary definitions, followed by an attachment statement wherein an object frame *instance* is instantiated). This is a very simple example as might be used for children's stories; for the sake of brevity it does not show the structural parent and "RelationshipToParent" infrastructure.

[24] Star includes a somewhat-skeletal taxonomy of attribute types that has two main categories at the top of the tree; these are referred to as "attribute super-types": they are the category of location attribute types and the category of value (qualitative) attribute types.

```
// Definitions

ObjectFrameClass VehicleObjectClass
{
   AttributeTypes
   (
      AttributeType "SpatialLocation"
      (
         <SuperType val = "Locational"/>

         "Values"
         (
            "Garage",
            "Driveway",
            "Roadway",
          );
      );

      AttributeType "Color"
      (
         <SuperType val = "Qualitative"/>

          "Values"
          (
             "Red",
             "Green",
             "Blue"
          );
      );
   );
};

// Attachments (Object Instantiations)

attach VehicleObjectClass Car1;

// Assertions

assert Car1::
  ( <Attribute ref = SpatialLocation
               val = "Driveway" />,
    <Attribute ref = Color
               val = "Blue" />
  );
```

The "assert" statement contains an expression that is the two-part attribute cluster: it can be interpreted as "the entity at the location called "Driveway" has a color value of "Blue". The essential features of the two-part attribute cluster are illustrated here: it contains at least one *locational* attribute that specifies the location of the object frame instance (the entity), and at least one *value* attribute that specifies the *infused* or *populated* value of the object frame instance.

## 4.15 Feature: The Collection

The collection is an abstraction that represents a set of object frame instances. The collection concept addresses the need for an implementation mechanism that corresponds to the universal quantifier of FOL.

## 4.16 Feature: Separation of Simple Assertions from Implicative Assertions

ROSS differentiates assertions into two main groups: those that involve implication (implicative assertions) and those that do not: these are referred to as simple assertions. Simple assertions correspond to a variety of "fact-like" representations, depending on the problem domain, and implicative assertions correspond to rules.

## 4.17 Feature: Rule Binder Construct

ROSS has a feature that is used in implicative assertions (rules) called the "binder".[25] A binder is a representational construct that explicitly specifies the locational relationship between all location entities in the antecedent and consequent of a rule.

## 4.18 Feature: Representation of Abstract Entities

ROSS can handle the representation of abstractions that involve entities that have physical attributes that are not relevant to a given domain. Examples of abstract entities can be found in many domains. From the legal realm these include the *law suit, tort, jurisdiction, legal action* and *will*. Examples from the area of finance include *income, expense* and *revenue*. The areas of human cognitive concepts and mental processes, human emotions and natural language communication include many others. It can be argued that other abstractions include the physics concepts of force, energy and potential.

The underlying perspective is that all entities – whether they are perceived and conceptualized as "concrete" or abstract are in fact concrete to the extent that they can be rooted in a spatial/temporal frame of reference. The Platonic *form* or *ideal* is not allowed existence in ROSS representations since all such supposed forms (e.g. a circle) only actually exist as concepts.

Representation of abstract entities involves *grounding* of such entities in a physical spatial/temporal frame of reference (for instance, within an NLU instance model, the frame of reference is a structural parent object frame instance). Two approaches are possible: the first approach involves the use of dimension systems that are *skeletal* (for instance, those that have less detail than a 3D Cartesian coordinate system). An example would be a dimension system that describes *legal jurisdiction* using an enumerated list of jurisdictional regions rather than more-specific geographical coordinates. The second approach happens during the assimilation of abstract entities into a structural parent frame of reference when they are instantiated in an instance model; during the attachment process they are *imprecisely situated* via the use of attribute value ranges.

[25] Hofford 2013 the U.S. patent "Expert System and method" introduced the *binder* concept.

An example involves the representation of revenue and is accomplished as follows. An instance of revenue (e.g. a sale of some item in a particular store) has at least some component in a four-dimensional space-time reality, but the coordinate system that is needed to describe it should be appropriate to the needs of the representation. This requires the use of a dimension system that has enumerated values that indicate time and place values. Secondly, the necessary attribute values are less-descriptive than those for physically concrete domains. For instance, the time at which a particular sales transaction took place may be specified as within the range of the business hours of the store where it was sold (e.g. TimeOfSale = range (9:00am – 6:00pm)). The exact place is not needed, and therefore a general place can be specified (e.g. PlaceOfSale = "MainStreetStore-399").

In many cases, an abstract entity is not one, but in fact an aggregation of multiple entities that exist at disparate places and times – a law suit would fit in this category. A ROSS fact repository can accommodate this type of aggregate entity – for instance an NLU instance model would contain specifications of various objects, processes, features, etc. of a law suit, using the standard set of ROSS primitives. However, entities that are involved in the law suit are also described as having *attributes* that are based on the fact that the humans that are involved have conceptualizations about such entities. E.g. the process of the *initial filing* of the law suit is performed (physically) by a person or persons, but it is also "known" by various persons as being a process that involves physical entities that are part of the larger abstraction of the law suit.

Entity classes that are defined primarily by their *behavior* are handled by the representational mechanism of the ROSS behavior class. In many situations, the spatial/temporal specifics are of minor importance relative to some essential functionality. An example is the abstraction of the *physical harm cause* that might be said to exist in relation to instances of a person class (this would be any thing – person or animal, falling rock, etc. that has a potential behavior involving "causes harm to a person"). A ROSS entity class would be defined (PhysicalHarmCausalAgent) that has a structural parent that is identical to the one used by a *threatened persons* class. In a fact repository (e.g. NLU instance model) an instantiated instance of PhysicalHarmCausalAgent is an object frame instance that exists at a point on a timeline, having a set of spatial relationship attributes in relation to the harmed person.

## 5 Star Language Definitions

This is an overview of the Star language syntactic features that are used for creating definitions. This section illustrates some of the main representation constructs using examples; other constructs are only briefly described[26]. (Note that comments are preceded by "//").

### 5.1 Constant Set Name Keywords

Star contains the following keywords that are names for "built-in" sets of constants:

- IntegerConstant
- FloatingPointConstant (although ROSS value sets are integer-based this is included for completeness)
- StringConstant

### 5.2 Built-in Attribute Super Types

There are two pre-defined attribute "super types", they are higher-level attribute type categories:

- Locational attribute types
- Qualitative attribute types

### 5.3 Built-in Attribute Value Set Super Types

There are also two pre-defined attribute value set "usage" super types, they correspond to the attribute type super types, and are:

- Locational attribute value set
- Qualitative attribute value set

### 5.4 Built-in Attribute Value Types

These are not "set types" – they are special categories for specific attribute values. They are:

[26] A parser and processor for the Star language does exist that implements most of the features described herein.

- SpaceValue
- NonSpaceValue

These categories play a special role in instance models and in inference.

### 5.5 The Dictionary Element

The version of Star that is used for NLU applications contains an element called "Dictionary" that can be used in a variety of contexts. This associates a set of words with a single concept. It has the capacity for multiple language support. In the following example, a Dictionary construct is used within an AttributeType statement in order to create a set of English words for each vehicle exterior color value.

```
AttributeType "VehicleExteriorColor"
(
  <SuperType val = "QualityAttributeType"/>

  "Values"
  (
    { "Black": Dictionary
        ( English
          ( { "black", "charcoal" } ); ); ,
      "Blue": Dictionary
        ( English
          ( { "blue" } ); ); ,
      "Silver": Dictionary
        ( English
          ( { "silver", "grey" } ); ); ,
      "White": Dictionary
        ( English
          ( { "white", "opal" } ); );
    }
  );
);
```

A word that is defined within a dictionary element is not limited to use in that element: for instance, the word "opal" in the example here may exist in any of a number of other places within other Dictionary elements.

### 5.6 Attribute Value Sets

An attribute value set is defined using the "ValueSet" keyword, followed by a value set name, and then by a value set expression. This example defines two value sets, "Millimeter" and "VehiclePhysicalDimension", which uses Millimeter. A declaration is also included here for the purpose of defining a constant value (the maximum length of a vehicle dimension).

```
ValueSet "Millimeter"
(
    IntegerConstant
);
Integer lenMaxVehiclePhysicalDimension =
                       12000;

ValueSet "VehiclePhysicalDimension"
(
    <BaseValueSet ref = Millimeter />
      // UnitOfMeasure
    <SuperTypeUsage val = "Locational" />
     { 1, .. lenMaxVehiclePhysicalDimension }
);
```

The attribute value sets defined here can subsequently be used in other statements and expressions as needed.

### 5.7 Mappings

Mapping statements are useful for mapping members of one value set to members of another value set. (In the example here, the value sets named "Meter" and "Foot" have already been defined).

```
Mapping "MeterToFoot"
(
     <Source ref = Meter />
     <Dest ref = Foot />
     <Function expr = (x$ * 3.2808) />
);
```

This can now be used by system processing components that need to convert between members of the value sets.

### 5.8 Attribute Types

The attribute type statement defines an attribute type. An example is as follows:

```
AttributeType "VehicleExteriorColor"
(
  <SuperType val = "Qualitative"/>

  "Values"
  (
    { "Black": Dictionary
        ( English
          ( { "black", "charcoal" } ); ); ,
      "Blue": Dictionary
        ( English
          ( { "blue" } ); ); ,
      "Silver": Dictionary
        ( English
          ( { "silver", "grey" } ); ); ,
      "White": Dictionary
        ( English
          ( { "white", "opal" } ); );
    }
  );
);
```

Once the attribute type has been defined, the defined attribute type name can then be used in other statements and expressions as needed. Where the attribute type is used, type checking can be performed for values that derive from the attribute value set.

### 5.9 Dimension System Types

The dimension system type definition (sometimes referred to as just "dimension system")

creates a dimension system (e.g. a coordinate system), that is used by object frame classes and instances. A dimension system is a means of aggregating attribute types that are intended for collective use into a group in order to fully describe the locational attributes of an object frame instance. (The expression that uses the attribute types in order to specify a specific set of attributes is referred to as a "dimension set expression"). For instance, a dimension system type for description of geographical positions would involve attribute type definitions for each of latitude and longitude. An example dimension system that represents Cartesian coordinates that use the millimeter as a grain size is as follows:

```
DimensionSystem "MillimeterCoordinates"
(
    SpatialAttributeTypes
    (
      "AttributeTypeX"
      (
        <SuperType val = "Locational"/>
        "ValueSet"
        (
          <BaseValueSet ref = Millimeter />
          <SuperTypeUsage val = " Locational" />
          { 1, .. lenMaxSpatialDimensionMillimeters}
        );
      );
      "AttributeTypeY"
      (
        <SuperType val = "Locational" />
        "ValueSet"
        (
          <BaseValueSet ref = Millimeter />
          <SuperTypeUsage val = "Locational" />
          { 1, .. lenMaxSpatialDimensionMillimeters }
        );
      );
      "AttributeTypeZ"
      (
        <SuperType val = "Locational"/>
        "ValueSet"
        (
          <BaseValueSet ref = Millimeter />
          <SuperTypeUsage val = "Locational" />
          { 1, .. lenMaxSpatialDimensionMillimeters }
        );
      );
    );
   // TemporalAttributeTypes
);
```

Note that this dimension system does not include an optional section for temporal attribute types.

### 5.10 Specification System Type

The specification system type incorporates a dimension system and an *inner content* section in order to create a system that can be used for fully specifying the place and qualitative value of unit-sized or aggregate object frame instances. The expression that uses a specification system is called a "specification set expression".

An example specification system that is useful for specifying the material composition of vehicle parts at the millimeter level of granularity is as follows:

```
SpecificationSystem "VehiclePartPhysicalComposition"
(
   DimensionSystem "VehiclePartCoordinates"
                    (MillimeterCoordinates);
   InnerContent
   (
      QualityAttributeTypes
      (
        "EssentialValueAttributeType"
          (MaterialCompositionAttributeType);
      );
   );
);
```

Note that the attribute type called "MaterialCompositionAttributeType" is not shown – this would consist of values such as "Aluminum", "Plastic", "Fabric", etc.

### 5.11 The Attribute

Attributes have already been described; attributes can exist within object frame classes, described next and they can exist within object frame instances within fact transcripts and instance models. An example of an attribute that belongs with a class would exist within a class for gold coins: all instances of this class can be said to have the attribute of compositionality of gold material.

### 5.12 Object Frame Class

The object frame class is the foundation for representation of the instances that get instantiated and thus exist in a fact transcript or in an NLU instance model. Object frame classes are also used within definitions (within other object frame classes) and in rules. The object frame class structure is shown below (this description includes some NLU-specific features).

```
ObjectFrameClass ->

   ObjectFrameClassName
   MassSubstance Boolean flag
   DictionaryPriorWord structure
     // (e.g. for "fire engine")
   Dictionary structure
   HigherClasses list
   StructuralParentClassesBase
   StructuralParentClassesTypicalImmediate
   RelationshipToParent structure
   AttributeTypes list
   Attributes list
   Templates (used for infusion)
   RelationshipTypes list
   DimensionSystems list
   Structure (list of ObjectFrameClass)
```

```
BehaviorClass list
```

The HigherClasses list represents all higher classes in the optional inheritance hierarchy for an object frame class. For instance, a Car class may get some of its attributes and structure via inheritance from a Vehicle class. The structural parent class items are lists that usually consist of a single item that represents the structural parent class of the object frame class. The RelationshipToParent structure contains attributes that specify how the object frame class is tied to structural parent classes. An example would involve a set of attributes relating an Engine class to a Car class.

The Structure item is where unchangeable sub-parts of the class are represented. The BehaviorClass list contains references to behavior classes that can be associated with object frame instances that are instantiated from the object frame class.

The following example object frame class illustrates some of the object frame class features: this class would model a simple (old-style) vehicle ignition key. (Most of the detail is not shown here, as indicated by placeholders).

```
ObjectFrameClass "IgnitionKeyObjectFrameClass"
(
  // placeholder:
  HigherClasses ();

  StructuralParentClassesBase
  (
    { "EverydayObjectStructuralParentClass" }
  );

  // placeholders:
  RelationshipToParent
  (
    AtLocations ();
    OrientationSpecifiers ();
    OuterDimensionSystemExtents ();
  );

  AttributeTypes
  (
    AttributeType "MaterialCompositionAttributeType"
    (
      <SuperType val = "Qualitative"/>
      "Values"
      (
         <SuperTypeUsage val = " Qualitative" />
          { "Steel", "Plastic", "Fabric"}
      );
    );
  );

  Attributes
  (
    // The presence here indicates that any instance of this
       class has this specific attribute:

    Attribute "MaterialComposition"
    (
      <Attribute ref = MaterialCompositionAttributeType
                 val = "Steel" />
    );
  );
  // placeholder:
  DimensionSystems ();

  // placeholder:
  Structure ();

); // IgnitionKeyObjectFrameClass
```

The potential behaviors of this class are not shown (this is explained in what follows). This class can be used to instantiate instances, e.g. in an NLU instance model. (For NLU applications, at the time of instantiation, a structural parent instance that derives from the "EverydayObjectStructuralParentClass" must exist or one will be automatically instantiated). It must be noted that the vehicle ignition key class shown here is dealt with only as a "unit" – for instance the "MaterialComposition" attribute is an attribute of the *whole*, not of its parts. The internal structure of such a class is dealt with using other features, as explained next.

## 5.13 Explanatory Note: How an Object Frame Class Implements Structure

Several of the member fields of the object frame class as shown above are used in coordination in order to represent nested physical structures. For instance, to model a vehicle class, a "vehicle" object frame class would be created; it has a set of dimension systems (using the DimensionSystems list). One or more such dimension systems provide a basis for the specification of the location of the sub-part object frame classes of the vehicle class. An engine compartment class is created and added as an item to the Structure list of the vehicle class. Unless its relative place within the vehicle class is specified, the engine compartment class can be said to "float" somewhere within the perimeters of the vehicle class. The RelationshipToParent structure of the engine compartment class specifies either specific location attributes of this component in relation to the parent engine class, or it can be used to declare placeholders that are used for such specification in fact repository artifacts when such information is available. These location attributes make use of the dimension system of the parent (the vehicle class), in order to specify both spatial place and spatial size (referred to as "extent").

An advantage of this approach is that it is effective for keeping track of real physical objects that have complex part-subpart structures.

### 5.14 Template Class

The template class can be understood using the metaphor of drawing: a template class describes a method that is used to draw a picture within an object frame range instance. A simple example would be a template class that contains a function to draw an oval within an object frame range instance that has a rectangular shape. A more complex example would involve a set of drawing instructions that can be used for drawing a face, or for the 3D rendering of a person's head within a cuboid-shaped object frame range instance. The process of drawing/rendering is referred to as "infusion".

### 5.15 Populated Object Class

The populated object class is a representational construct that allows for the application of a coordinated set of values to an aggregate object frame class. The process of applying a populated object class to an object frame class is referred to as "population". Populated object classes are used within behavior classes, described next.

### 5.16 Behavior Class

The behavior class is the basis for describing behaviors. A behavior class associates a set of "prior" states with a set of "post" states. Examples of behavior classes for the PersonObjectFrameClass class include "PersonWalks" and "PersonCommunicates". Behavior classes have the following structure:

- A bridge structural parent class – a reference to an object frame class that contains a dimension system that must be shared by all object frame classes in the behavior class, so that locational relationships can be specified within the binder construct that ties objects of the prior states section to objects of the post states section.
- A PriorStates section, consisting of a list of populated object classes. This is like the antecedent (the "if part") within a rule.
- A PostStates section, consisting of a list of populated object classes. This is like the consequent (the "then part") within a rule.

### 5.17 Explanatory Note: How Behaviors are Related To Object Frame Classes and Instances

The object frame class and the behavior class have both been described, but the question of how they are related has not been addressed. It must be noted that an object frame class does not actually have behaviors – it only has a list of *potential* behaviors. An object frame instance does not have behaviors at all: it only implements states of a behavior at some point along a timeline. For example, within an NLU instance model, each single-time-point object frame instance participates in behaviors via attributes that specify its state. An aggregate object frame instance (at a single time point) can thus participate in multiple behaviors simultaneously due to its having multiple attributes, each of which represent some aspect of its state.

## 6 Knowledge Base Concepts

This section deals with concepts that pertain to ROSS knowledge bases.

### 6.1 Knowledge Bases

A ROSS knowledge base, or Infopedia, contains Star language definitions (and optionally, rules). It includes a mixture of definitions that cross the spectrum from universal and generic to domain-specific. There are generic object frame classes for high-level abstract objects, and more-immediate object frame classes such as *PersonClass* and *VehicleClass*. The Infopedia also contains a variety of supporting definitions for attribute value sets, attribute types, value set mappings, and dimension system types.

### 6.2 Knowledge Acquisition/Ontology Derivation

Learning of classes is an important area that can make use of the features of ROSS. The use of learning techniques is not an absolute necessity since both generic and domain-specific ROSS definitions can be created by a human knowledge engineer or ontology practitioner. (It is more ideal for some ROSS definitions at the higher levels of abstraction to be hand-crafted in advance rather than learned – e.g. attribute value sets, attribute types and dimension systems). Since knowledge engineering has long been recognized as a bottleneck for AI, automated approaches are viewed as highly valuable. The following are several broad categories of automated knowledge acquisition to learn classes and class features from natural language text:

- Intermediate-depth approaches that learn features based on associations. E.g. (unsupervised) learning that cars can be blue based on sentences that associate "blue" with "car".
- Learning new sub-classes and their behaviors based on simple sentences of the form "an x is a y that does z". (E.g. "an electrician is a person who fixes electrical problems").
- Deeper approaches that learn structure, features and behaviors from NL descriptions that explicitly describe structure and features. (E.g. a "simple encyclopedia" entry on "automobile").

The Ontology derivation task is not limited to natural language-based approaches. Other possibilities include the use of interactive tools that such as those that would allow human users to draw objects. Another approach would involve the processing of engineered specifications to generate ROSS classes.

## 7 Fact Repository Concepts

This section deals with concepts that pertain to ROSS fact repositories and with the processes such as instantiation that generate the information that exists in fact repositories.

### 7.1 Fact Repositories: Transcripts and Instance Models

There are a variety of representational artifacts that use the ROSS approach with representational constructs that are *fact-like*. The term "fact-like" includes representations that are true "facts" about past situations, and it includes other assertions such as plan goals and predictions. "Fact repository" is defined to include any representational artifact containing such constructs. A fact repository has a top-level structure: the repository may represent multiple situations, e.g. situations that are a mixture of ones from the past (from various places and times), others that are present-tense, and some that are hypothetical.

### 7.2 Transcript Types

A transcript is a document that contains fact-like representational constructs for use in AI automated reasoning applications. There are a number of transcript types that use ROSS. These include the following:

- Past fact transcripts that are useful for automated reasoning about past fact situations (e.g. fault diagnosis)
- Specification transcripts for automated inference for design or planning problems; these transcripts contain fact-like constructs that include predicted states and goals

### 7.3 Instance Model

An *instance model* is a type of fact repository for NLU: it is a meaning representation instance that represents factual information about past and/or present situations and events. It is an artifact that is a structured representation of the subject matter of an input natural language text fragment such as a story.

An instance model contains a list of *contexts*. The order of contexts in the list usually corresponds to the order of occurrence of sentences in the input text. (The appendix contains a partial instance model).

### 7.4 NLU Instance Model Context

An instance model *context* is a representational construct that pertains to a particular space and time frame of reference. (The structure that is contained within a single context is analogous to a diorama). The discourse of a story may have many such contexts. For instance, a story may contain the following two sentences in sequence: "A Seattle home was burglarized late yesterday. John Smith owns the home". The first sentence is in the past tense and is the basis for a context. The second sentence is in the present tense and thus provides the basis for a second and separate context. An instance model contains at least one context.

The context concept may also be used to represent the content of spoken or written communication. In this case it designates a separate frame of reference that represents information that was communicated by a human agent.

An instance model context represents a single *situation*, described next.

### 7.5 The Fact Repository Situation

A fact repository may contain one or multiple *situations*. A *situation* is a collection of related facts, each of which involves entities that all share a common structural parent instance.

## 7.6 The Structural Parent Class

Structural parent object frame *classes* exist in knowledge bases as definitions; structural parent object *instances* exist in fact repositories, e.g. in instance models.

## 7.7 The Structural Parent Instance

A situation contains a single top-level object frame instance that serves a special function as a "structural parent". The structural parent object frame instance has an *InstanceStructure* section that specifies all object frame instances that are immediate children that are within the spatial and temporal range of the structural parent object frame instance.

## 7.8 Object Instances

The structure of an object frame instance is shown here.

```
ObjectInstance ->

   ObjectFrameClassName
   ObjectInstanceUniqueIdentifier
   CausalityRole
   PersonOrPlaceIndicator
   RelationshipToParent structure
   Attributes list
   Relationships list
   InstanceStructure  (structure  containing
list of object instances)
```

Object instances are instantiated using object frame classes – thus the first field of an object instance is the object frame class from which it was instantiated. The next field is a unique identifier that refers to the instance as it exists or existed in the space-time frame of reference of the structural parent of the context.

The next field, CausalityRole, designates whether the object instance is part of a cause or part of an effect. (NLU-specific: the PersonOrPlace indicator represents information determined by a semantic engine that is used in creating bulleted summaries). If the object instance is the structural child of a parent object instance, the RelationshipToParent structure specifies the specific attributes that relate the child to the parent. The Attribute and Relationship lists contain attribute and relationship attribute information about the object instance. Finally, the InstanceStructure is a collection of references to all child instances. For instance, the representation of a "car" instance would typically contain object instances here for "engine", "transmission", "body frame", etc.

## 7.9 Instantiation

Instantiation is the process of creating an object frame instance within a fact repository from an object frame class; it involves the sub-tasks of *attachment* and of *infusion* or *population*. (For purposes of illustration, each of these concepts is described here in terms how it is performed by a NLU semantic engine, e.g. when the engine generates an object instance within an instance model).

## 7.10 Attachment

*Attachment* is the process of creating an object frame instance. When a structural parent object frame instance is created within a situation, it is simply given a unique identifier or name. However when an object frame instance that is a child of a structural parent, or of other object frame instances is created, attachment involves creating the instance, giving it an identifier or name, and then setting its RelationshipToParent attributes. It also involves specifying a reference to the child instance within the InstanceStructure section of the parent instance.

The effect of performing a group of attachments can be visualized as analogous to a process of creating a diorama frame and then inserting various empty smaller wire-frame structures (some nested within others) into it.

## 7.11 Infusion and Population

The process of *infusion* operates on empty object frame instances: it sets actual values for them. Infusion as applied to a unit-sized object frame instance just involves setting its value. Infusion of a value into an object frame range instance makes use of a template class. *Population* is similar to infusion and involves using a populated object class to set the values of an aggregate object frame instance.

## 7.12 Permanence and Perpetuation

Practical considerations of creation and maintenance of representations in a fact repository artifact may necessitate the use of convenience assumptions (described in the above Ontology section). The first of these is the *empty space* assumption: within a structural parent (similar to a diorama), at the first time point and for all subsequent time points, any unit-sized location that has not been specifically and overtly infused or populated is assumed to have a value that inherits from the SpaceValue value category.

The perpetuation assumption involves perpetuation along a time line; it can be used in similar fashion as the empty space assumption: the assumption is that for any unit-sized location that has been infused with a value at time t=n, it can be assumed that the subsequent unit-sized location at the same spatial location (at time t=(n+1)) will have the same value unless it is overtly specified to have a different value. This assumption is useful for stationary objects but does not address the representation of objects in motion.

## 8 Automation of Inference

The representational features of ROSS provide a foundation for automated reasoning that is open-ended and unrestricted due to the loose coupling between representation and reasoning. The following are a few examples of broad categories of reasoning tasks that can be accomplished using the ROSS method:

- Reasoning about situations and events that occurred in the past to perform *past fact derivation*. This category includes diagnosis of faults/failures.
- Reasoning from requirements specifications to generate design artifacts.
- Reasoning from plan goals to generate plans.

Detailed information on the structure of rules and about the reasoning algorithms of the expert system for diagnosis that was developed by the author is available in Hofford (2013).

## 9 NLU: Semantic Processing

ROSS has capabilities that are important for NLU semantic analysis and processing. A semantic engine that is part of a modular NLU system (one that separates the syntactic processing stage from the semantic processing stage) is briefly described. The area of *story comprehension* – a restricted subset of the very broad range of NLU tasks - is used for purposes of demonstrating the operation of such a system as it uses and produces ROSS-based information.

### 9.1 Data Flow

The *semantic engine* of such a system requires a set of Star language definitions as input. These definitions include supporting definitions, object frame classes and behavior classes. The Star language definitions are stored in text files and are read in at the time of system initialization and used to build an in-memory knowledge base (the Infopedia).

The second main input to the engine is a list of syntax trees that contain semantic role and other information. Each syntax tree in the list represents a single sentence; sentences may be nested within sentences.

The main output of the engine is an instance model.

### 9.2 Class Selection Algorithm

An important task for the engine is that of class selection. For instance, the input may describe a "car". ROSS is not limited to a single "car class" – there may be any of a number of classes that represent "car", each of which emphasizes some set of aspects of a car. For instance, employees in an automobile manufacturing plant would have a concept of car that differs in many ways from the concept of car that is used by consumers who are drivers. Each car class would associate the word "car" with itself using the Star *Dictionary* element. The task of disambiguation in order to select the most appropriate class is potentially open-ended; a variety of heuristics and inference techniques may be used.

### 9.3 Integration of the Syntactic and Semantic Phases

Automated reasoning by the NLU engine in order to provide feedback to the natural language parser for purposes of disambiguation is an area of ongoing research.

## 10 Conclusion

In comparison with logic and other AI representational schemes, the essence of the ROSS approach centers on its *ontological restrictions*, or *restrictions for modeling*. This ontological framework is rooted in a naïve view that uses discreteness of space and time as a basis. The benefits are very substantial in how it enables modeling and representations that are very rich and that are appropriate in semantic complexity for the comprehensive and well-organized representation of domains. The ROSS ontological restrictions and commitments provide a set of guidelines for the task of *knowledge base* creation and refinement. Furthermore, these restrictions facilitate the creation of *fact representations* that are highly structured, unambiguous and non-redundant. The ROSS framework for representing *physical structure* – using dimension systems, the structural features of the object frame class, and the structural parent instance concept - is beneficial for providing a frame of reference that provides an organizing infrastructure for the representation of relational attributes (the normal form). ROSS instance representations are *implicitly hybrid*, combining an analogical approach with a symbolic approach. The analogical aspect is present due to the use of a structural parent instance (as it houses a base/master dimension system). When this is used as a foundation for ROSS *normal form* information (e.g. instantiations of shape templates), the resultant fact instance representations provide for query (as against structured information) and are a platform for inference.

The ROSS approach to class hierarchies is bottom-up rather than top-down. The *instantiation* process that uses classes to create instances (e.g. in NLU instance models) starts from the perspective of assuming too little, not too much. For instance, during the NLU semantic processing phase, few details may be known about an instance of a class represented by a term in the input text; therefore ROSS starts with fewer assumptions about its class instance features, structure and attributes.

Secondary information can be derived from primary information and stored in the same fact repository. ROSS normal form provides a foundation; the infrastructure (e.g. of an instance model) is flexible so that additional information such as relationships can be stored in the same artifact. Derived information such as collections also fits into the same representational artifact.

The richness of the ROSS approach has implications for the representation of *commonsense knowledge* – this has value for areas such as NLU story comprehension.

### 10.1 Ongoing Research

A number of challenges have provided motivation for further research; they include:

- Coordination of the ROSS discrete model with representations of continuous phenomena.
- Ontology derivation/knowledge acquisition.
- Integration of probabilistic information, e.g. classes with fuzzy boundaries.
- Integration of procedural "sub-symbolic" connectionist approaches (e.g. neural networks) with ROSS.

## Appendix: XML Instance Model (showing partial detail)

{ Notes:
- this is a fragment of an NLU instance model that contain an instance of a house cat (it is part of an instance model that can represent the sentence "A cat is on a mat.")
- the class names, e.g. HouseCatObjectFrameClass would have been defined in a knowledge base
- this demonstrates the use of a timeline within a structural parent instance

}

```
<?xml version="1.0" encoding="us-ascii" standalone="yes"?>

<InstanceModel>

   <!—NOT SHOWN:  Source document information section -->

  <ConceptualModel>
    <GlobalAssumptions>
      <!-- Any location that has not been infused has a value that
           inherits from the "EmptySpace" value category -->
      <EmptySpaceAssumption value ="true" />
      <!-- Attached objects are permanent through time -->
      <PermanentAttachmentsAssumption value = "true" />
      <!-- Stationary values at t = n perpetuate forward in time -->
      <PerpetuationAssumption value ="true" />
    </GlobalAssumptions>

    <!-- The following element attaches the main structural parent instance.
         This instance is analogous to a diorama. -->

    <StructuralParent class="EverydayObjectStructuralParentClass"
              name="EverydayObjectStructuralParentInstance-1"
              type="Range"
              timeline="EverydayObjectStructuralParentClass.EverydayObjectBasicDimensionSystem">

      <TimePoint value = "T01">

        <!-- The InstanceStructure element contains attachments, attributes, and template-
             based infusions for each of the main components within the main structural
             parent instance. Each object frame instance, e.g. the cat, is analogous
             to an empty wire frame within the diorama.-->

        <InstanceStructure>

          <Component class="HouseCatObjectFrameClass"
                    name="CatObjectFrameClass-Instance1"
                    type="Aggregate">   <!— Aggregate indicates it has components -->

            <!—NOT SHOWN:
                - RelationshipToParent attributes (AtLocations, OrientationSpecifiers, Extents)
                - Infused values
                      - Attributes of the whole, e.g. Height = 225 mm, e.g. FurColor = grey
                - Relationships of this instance, e.g. the distance to the mat
                - InstanceStructure, containing components, e.g. the cat's head and body
                    -> Components such as cat's head may use templates or shape templates for infusions
                       so that the 3D region has a set of populated values (representing a cat's head)
            -->
          </Component>

        </InstanceStructure>
      </TimePoint>
    </StructuralParent>
  </ConceptualModel>
</InstanceModel>
```